\documentclass{article}
\usepackage{ijcai24}

\usepackage{times}
\usepackage{soul}
\usepackage{url}
\usepackage[colorlinks=true,hyperfootnotes=false]{hyperref}
\usepackage[utf8]{inputenc}
\usepackage[small]{caption}
\usepackage{graphicx}
\usepackage{amsmath}
\usepackage{amsthm}
\usepackage{booktabs}
\usepackage{algorithm}
\usepackage{algorithmic}
\usepackage[switch]{lineno}
\usepackage{subcaption}

\usepackage[T1]{fontenc}
\usepackage{textgreek}
\usepackage{amsfonts}
\usepackage{color}
\usepackage[dvipsnames]{xcolor}
\usepackage{longtable}
\usepackage{tabularx}
\usepackage{multirow}
\usepackage{bm}
\usepackage{enumitem}
\usepackage{listings} % Add this in your preamble
\usepackage{fancyvrb}
\usepackage[most]{tcolorbox}
\newcommand{\unnumberedfootnote}[1]{%
    {\let\thefootnote\relax\footnotetext{#1}}%
}

\definecolor{RoseRed}{RGB}{245, 102, 98}
\definecolor{MyBlue}{RGB}{125, 164, 247}
\definecolor{MyGreen}{RGB}{152, 211, 90}
\definecolor{MyOrange}{RGB}{244, 177, 131}
\definecolor{LightBlue}{RGB}{173, 216, 230}

\title{Prompting Large Language Models for Zero-Shot Clinical Prediction with Structured Longitudinal Electronic Health Record Data}

\author{
Yinghao Zhu$^{1,2~\ast~\ddagger}$
\and
Zixiang Wang$^{2~\ast}$\and
Junyi Gao$^{3,4~\ddagger}$\and
Yuning Tong$^{5}$\and
Jingkun An$^{1}$\and\\
Weibin Liao$^{2}$\and
Ewen M. Harrison$^{3}$\and
Liantao Ma$^{2~\dagger}$\and
Chengwei Pan$^{1~\dagger}$
\\
\affiliations
$^1$Institute of Artificial Intelligence, Beihang University, Beijing, China\\
$^2$National Engineering Research Center for Software Engineering, Peking University, Beijing, China\\
$^3$Centre for Medical Informatics, University of Edinburgh, Edinburgh, UK\\
$^4$Health Data Research UK, London, UK\\
$^5$School of Computing, National University of Singapore, Singapore\\
\emails
\{zhuyinghao,pancw\}@buaa.edu.cn, malt@pku.edu.cn, junyii.gao@gmail.com
}

\begin{document}

\maketitle
\unnumberedfootnote{$^\ast$ Equal contribution, $^\dagger$ Corresponding author, $^\ddagger$ Project leader.}

% \title{Assessing the Zero-Shot Capacities of Large Language Models for Clinical Prediction with Structured Longitudinal EHR Data}

% PromptCast: A New Prompt-based Learning Paradigm for Time Series Forecasting
% evaluation & prompt -> understand for LLM
% A Prompt-based Framework for Ze

% Prompting Large Language Models for Zero-Shot Clinical Prediction with Structured Longitudinal EHR Data
% Prompting Large Language Models for Zero-Shot Clinical Prediction with Structured Longitudinal Electronic Health Record Data

\begin{abstract}

The inherent complexity of structured longitudinal Electronic Health Records (EHR) data poses a significant challenge when integrated with Large Language Models (LLMs), which are traditionally tailored for natural language processing. Motivated by the urgent need for swift decision-making during new disease outbreaks, where traditional predictive models often fail due to a lack of historical data, this research investigates the adaptability of LLMs, like GPT-4, to EHR data. We particularly focus on their zero-shot capabilities, which enable them to make predictions in scenarios in which they haven't been explicitly trained. In response to the longitudinal, sparse, and knowledge-infused nature of EHR data, our prompting approach involves taking into account specific EHR characteristics such as units and reference ranges, and employing an in-context learning strategy that aligns with clinical contexts. Our comprehensive experiments on the MIMIC-IV and TJH datasets demonstrate that with our elaborately designed prompting framework, LLMs can improve prediction performance in key tasks such as mortality, length-of-stay, and 30-day readmission by about 35\%, surpassing ML models in few-shot settings. Our research underscores the potential of LLMs in enhancing clinical decision-making, especially in urgent healthcare situations like the outbreak of emerging diseases with no labeled data. The code is publicly available at~\url{https://github.com/yhzhu99/llm4healthcare} for reproducibility.
% The code is publicly available at~\url{https://anonymous.4open.science/r/llm4healthcare} for reproducibility.

% \yh{TODO: improvement score}
% \jy{Emphasize the challenge and contribution in prompt designing}
% \yh{TODO: code open source}
\end{abstract}

\section{Introduction}
% \cw{Emphasizing zero-shot prediction in EHR is very important both in practical clinical scenarios and research. Continuously state the significance of the work of the paper.}
% \yh{Emphasize difference with traditional text-based data (tabular data); Prediction task level (Not QA, diverse clinical prediction tasks -> then discuss temporal requirements, and multitask)}

\begin{figure}[!ht]
    \centering
    \includegraphics[width=1.0\linewidth]{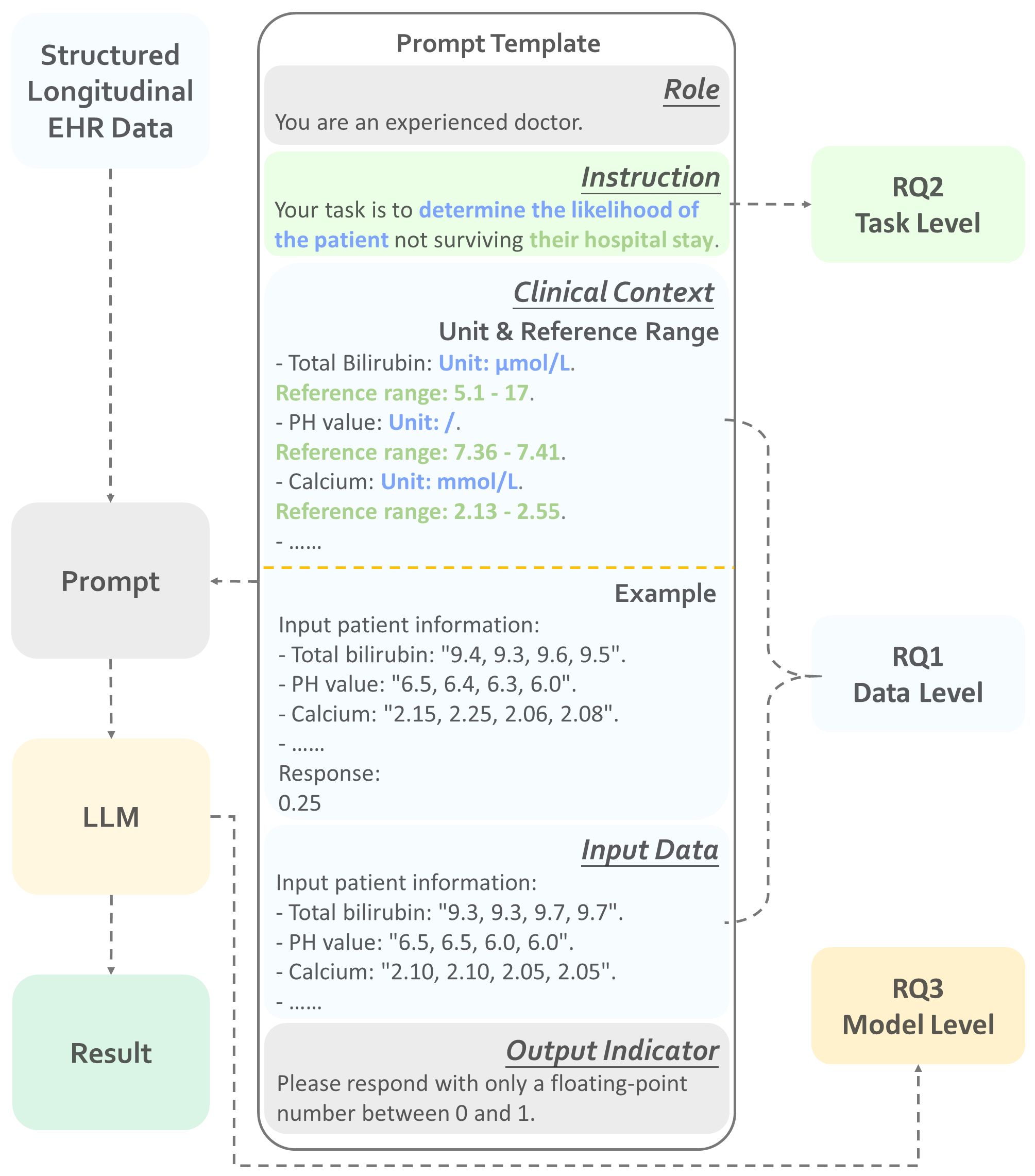}
    \caption{\textit{Proposed prompt template which incorporates five key elements in prompt engineering: role, instruction, clinical context, input data, and output indicator \& Overall structure of the paper.}}
    % \yh{Align with Data, Task, Model three aspects}
    % \jy{text too small and text color too light}
    \label{fig:prompt_template}
\end{figure}
% \cw{The color is too light to see clearly}
% \yh{Add an example figure of what composed of EHR data}
% \yh{Add figure of our naive prompting approach}
% \yh{TODO: 5 key roles in LLM: role, instruction, context, input data and output indicators, directly explain in prompt template figure}

% Electronic health records (EHRs) are pivotal in healthcare, providing critical insights for patient care. \jy{Do not talk about our research here}Our research emphasizes structured longitudinal EHR data which are predominantly utilized in clinical settings, offering a more privacy-conscious alternative to clinical notes~\cite{keshta2021security}. These records enable the prediction of patient outcomes and disease progression, greatly improving patient care and treatment planning. However, in emergencies like new diseases where historical data is scarce, the importance of EHR data becomes even more crucial~\cite{gao2024comprehensive,stephens2009using}\jy{logic gap here. the new disease does not make ehr more important.}. Traditional machine learning models, which are typically effective, require extensive data for accurate predictions~\cite{wang2020generalizing}. This presents a challenge in emergent diseases, known as a `cold start' problem, due to the limited data and incomplete understanding of vital diagnostic features~\cite{lauer2020incubation,ma2021distcare}.

Electronic health records (EHRs) are pivotal in healthcare, providing critical insights for patient care and enabling the prediction of patient outcomes and disease progression, greatly improving patient care and treatment planning. However, when meeting emergencies like new diseases where historical data is scarce~\cite{gao2024comprehensive,stephens2009using}, traditional machine learning models which require extensive data for accurate predictions struggle~\cite{wang2020generalizing}. Such scenarios present a `cold start' challenge due to the limited data~\cite{lauer2020incubation,ma2021distcare}. If zero-shot clinical prediction can be achieved, it would not only substantially enhance patient care in real-time clinical settings but also advance the frontiers of medical research, opening new avenues for understanding and managing emergent health challenges~\cite{wornow2023shaky}.

% and incomplete understanding of vital diagnostic features~\cite{lauer2020incubation,ma2021distcare}.

Large Language Models (LLMs) like GPT-4 are celebrated for their remarkable zero-shot learning abilities across various tasks involving unstructured text data, as evidenced by their performance in language translation, summarization, mathematical problem-solving, and code generation, etc~\cite{openai2023gpt4,touvron2023llama2}. Encouraged by this, we hypothesized that LLMs could be adept at performing zero-shot predictions on complex EHR tasks using structured longitudinal EHR data, such as mortality prediction, by leveraging their extensive medical knowledge acquired during training or inferencing~\cite{openai2023gpt4,touvron2023llama2}. However, applying LLMs to EHR data under zero-shot settings is nontrivial. We have proposed three-level research questions (RQs) to identify the fundamental disparities between clinical prediction and traditional natural language processing settings:  % RQ-1

\begin{itemize}[leftmargin=*]
    \item \textbf{RQ1 (Data-Level) What constitutes an effective EHR data prompt for LLMs?} From the data level, structured EHR data, characterized by its longitudinal nature, sparsity, and infusion with domain-specific knowledge, differs from the unstructured text or code data on which LLMs are trained~\cite{yang2024code_into_gpt}. The longitudinal aspect demands prompts that effectively represent time series data, the sparsity requires the LLM to be cognizant of missing information, and the knowledge-infused nature necessitates the integration of reliable medical knowledge context. Given our initial ineffectiveness of naive prompting strategies, how can we enable LLMs to comprehend EHR data, which primarily consists of numerical time-series values, a format that LLMs are not inherently skilled in handling~\cite{jin2021numgpt}? Such challenges necessitate more sophisticated prompt engineering to effectively bridge this gap at the data level~\cite{chacko2023paradigm}.   
    % \jy{Explain why this is an issue for prompt construction} % Given the initial ineffectiveness of naive prompting strategies, how can we enable LLMs to understand EHR data beyond vanilla numerical values?
    \item \textbf{RQ2 (Task-Level) Are LLMs capable of diverse clinical prediction tasks with various time spans?}
    % \yh{Observation: Temporal Contexts}
    Unlike medical Q\&A tasks focused on text-based symptom descriptions, EHR data covers diverse prediction tasks like mortality, length-of-stay, and readmission, presenting a distinct challenge for LLMs in processing various clinical predictions. Crucially, LLMs must effectively discern and predict outcomes over varying time spans in clinical practice, from immediate health outcomes to long-term projections.
    % instructions that demand detailed prediction objectives
    % sensitive to temporal changes, such as differentiating between immediate outcomes at discharge and long-term projections.
    % \jy{explain what are temporal changes}
    % RQ-2% Given the multifaceted nature of healthcare analytics, it is imperative to assess whether LLMs can maintain their versatility and accuracy across a range of predictive tasks, such as mortality prediction, length-of-stay, and readmission. Moreover, considering the varying temporal contexts from ICU to chronic disease scenarios, it is essential to understand the sensitivity of LLMs to time.
    \item \textbf{RQ3 (Model-Level) What are the zero-shot performance differences of various LLMs in handling EHR data, and can they outperform traditional machine learning models in few-shot settings?} At the model level, facing a diverse range of both open-sourced and closed-sourced LLMs, it is essential to determine, using our prompting framework that incorporates clinical contexts, the extent of improvement achieved. This includes identifying which LLMs excel in zero-shot scenarios and whether they can surpass traditional ML/DL models in few-shot settings. Such insights are crucial for clinicians when selecting the most suitable model for emerging diseases.
    % Meanwhile, at the model level, confronting a diverse landscape of both open-sourced and closed-sourced LLMs, it is crucial to determine with our prompting framework that incorporates clinical contexts, how much improvement is gained, which LLMs perform best in zero-shot scenarios and whether they can surpass traditional ML/DL models in few-shot settings. Such insights are invaluable for clinicians in selecting the most suitable model while facing emerging diseases. % This comparative analysis is essential to benchmark the capacities of LLMs in handling EHR data, offering valuable insights for future advancements in the development of LLMs.
\end{itemize}
    % \yh{Del RQ3, serve as contribution}

In this work, we aim to answer these research questions with a well-designed prompting framework while considering EHR data characteristics. We conduct comprehensive experiments on two real-world datasets, TJH~\cite{tjh} (a COVID-19 dataset collected in February 2020 to simulate emerging disease scenarios) and MIMIC-IV~\cite{mimic4} (simulating ICU scenarios). \textbf{For RQ1}, we explore the roles of longitudinality, sparsity and contextual learning strategies in enhancing the interpretability and accuracy of LLMs' EHR data analysis. \textbf{For RQ2}, we investigate the effectiveness of LLMs across a range of critical healthcare predictive tasks. We also assess the LLMs' sensitivity and adaptability to changes in temporal contexts in task instructions. \textbf{For RQ3}, we conduct a comprehensive comparison of four LLMs in a zero-shot setting with competitive ML and DL models in a few-shot setting for the TJH and MIMIC-IV mortality prediction tasks.
% \jy{summarize the model you chose. we conduct a comprehensive comparison among x llms. }.
Figure~\ref{fig:prompt_template} shows the proposed prompt template and overall structure of the paper.

% \yh{Only reserve the method part, move results to conclusion section}

% To the best of our knowledge, this is the first paper to discuss and benchmark in detail the prompting strategies for EHR data. We underscore that \textit{with our elaborately designed prompting framework, LLMs particularly GPT-4 can understand and perform zero-shot predictions for diverse healthcare tasks using structured longitudinal EHR data.} We improve 35.10\% and 25.43\% AUROC performance, 32.93\% and 63.92\% AUPRC performance on TJH and MIMIC-IV mortality prediction task than basic prompts. In summary, our research navigates the prompting strategies of LLMs for structured longitudinal EHR data in zero-shot clinical predictions, offering a path forward in leveraging these LLMs for enhanced healthcare predictions in the context of emerging diseases.
% \yh{TODO: Emphasize our contribution (our designed prompting framework), quantitive analysis, improve by xxx\%}
% Our findings provide insights into the adaptability of LLMs across various healthcare prediction tasks, contributing to a nuanced understanding of their application in healthcare.
% \yh{TODO: PRC improvement}

The remainder of the paper is structured as follows: Section~\ref{sec:related_work} reviews the related work. Section~\ref{sec:experimental_setups} details our experimental setups including problem formulation, datasets, and models. Sections~\ref{sec:rq1}, \ref{sec:rq2}, and~\ref{sec:rq3} present the three research questions with their motivations, methodologies, and results, respectively. Section~\ref{sec:discussions} discusses limitations and future work. Section~\ref{sec:conclusions} concludes the paper.
% \jy{Add a para about the structure of the following contents. In Section 2, we discuss.... In section 3, we discuss...}
% \yh{TODO: section reference}

% Figure~\ref{fig:prompt_template} shows our designed specialized prompt for clinical prediction which incorporates five key elements in prompt engineering: role, instruction, clinical context, input data, and output indicators~\cite{openai_prompt_engineering}.

\section{Related Work}\label{sec:related_work}

% \jy{If you didn't compare them, don't discuss them. Simply mention the line of few-shot learning works and emphasize why your work is different from their settings.}
% \jy{remove this section to appendix if no space}
Structured longitudinal EHR data is essential in clinical decision-making, offering a detailed history of patient healthcare interactions~\cite{gao2020dr,gao2020stagenet,ma2020adacare,ma2020concare,zhang2022m3care,zhu2023leveraging,ma2023aicare,zhu2023m3fair}. However, emerging diseases pose a significant challenge due to the scarcity of labeled EHR data in these contexts. To solve this issue, traditional methods have primarily relied on transfer learning~\cite{sun2019general,ma2021distcare,zhang2023transfer} and meta-learning~\cite{zhang2019metapred,chang2023meta}. While successful in transferring knowledge, these methods are complex, requiring the training of different models across multiple datasets or tasks. They also face limitations in understanding unaligned feature names and prove ineffective in scenarios lacking training labels.

Recently, large language models like ChatGPT (GPT-3.5), GPT-4, Gemini, and Llama 2 have demonstrated impressive zero-shot learning capacities on unseen tasks~\cite{blog2023ChatGPT,openai2023gpt4,pichai2023gemini,touvron2023llama2}. Trained on extensive text corpora, including medical data, they show proficiency in medical Q\&A tasks~\cite{nori2023GPTcapabilitiesOnMed}. However, these models struggle with numerical reasoning~\cite{jin2021numgpt}, a common requirement in clinical settings, and are not inherently equipped to handle structured longitudinal EHR data. Recent efforts have aimed to adapt LLMs to EHR data. Xue and Salim integrate LLM embeddings with external trainable projection layers~\cite{xue2023promptcast}. Prompt tuning approaches have also been investigated~\cite{liu2023large}. Concurrently, research into building EHR-specific foundation models through pretraining on EHR datasets, focusing on predicting the next ICD code, has been conducted~\cite{guo2023ehr,wornow2023shaky,wornow2023ehrshot}. These approaches focus on grounding the structured longitudinal EHR data to LLMs. However, they still face limitations in scenarios with an extreme lack of labels, as they require training or prompts on a few shots. The zero-shot capacity of LLMs on EHR data remains unexplored.

\section{Experimental Setups}\label{sec:experimental_setups}

\subsection{Problem Formulation}

%Structured longitudinal EHR data consist of a sequence of dynamic information such as lab tests and vital signs, as well as static information like demographics (e.g., sex and age) for each patient across multiple visits. Our objective is to encode this EHR data into a format that LLMs can process and 
In this work, we evaluate the zero-shot prediction ability of LLMs on various clinical prediction tasks including:
% To achieve this, we utilize pre-designed prompt templates that primarily encapsulate the raw EHR data, with the option to include additional external information, such as system prompt~\cite{zheng2023SystemPrompt}, units or reference ranges, to potentially enhance the LLMs' comprehension.

% We approach the analysis as a series of predictive tasks including regression and classification, each with its own characteristics:

\begin{itemize}[leftmargin=*]
    \item \textbf{Binary Classification Tasks:} \textit{(in-hospital mortality prediction and 30-day readmission prediction)} Predicting patient mortality outcomes ($0$: alive, $1$: dead) and readmission likelihood ($0$: no readmission, $1$: possible readmission). Here, we use AUROC, AUPRC, and min(+P, Se)~\cite{ma2022safari} metrics to evaluate the tasks' performance.
    % \item \textbf{Regression Task:} \textit{(length-of-stay prediction)} Estimating the length of hospital stay for patients. \jy{regression}
    \item \textbf{Regression Task:} \textit{(length-of-stay prediction)} This task involves estimating the continuous variable of a patient's hospital stay duration. We adopt MAE, MSE and RMSE as regression metrics. 
    \item \textbf{Multi-task Setting:} Simultaneous predictions of multiple outcomes (i.e., mortality and readmission).
\end{itemize}

Additionally, considering the multiple visits that a patient may have, the tasks vary in their temporal context. For instance, in the length-of-stay prediction, we predict at each visit, whereas for mortality and readmission tasks, a single prediction is made, aggregating all visits as input. This approach allows us to evaluate both the episodic and cumulative nature of a patient's medical history. Furthermore, we have instructed the LLM to respond with ``I do not know'' in cases where it is unable to provide a reasonable conclusion. In such instances, we assign a logit value of 0.50 for binary classification tasks, indicating a high level of uncertainty.

\subsection{Datasets}

% This study utilizes two primary datasets: TJH~\cite{tjh} and MIMIC-IV~\cite{mimic4}:

\begin{itemize}[leftmargin=*]
    \item \textbf{TJH Dataset}~\cite{tjh}: Derived from Tongji Hospital of Tongji Medical College, the TJH dataset consists of 485 anonymized COVID-19 inpatients treated in Wuhan, China, from January 10 to February 24, 2020. It includes 73 lab test features and 2 demographic features. This dataset is chosen for its relevance to the early stages of the COVID-19 pandemic, a period characterized by a sudden surge in cases and significant pressure on healthcare services. As such, the TJH dataset serves as an ideal representation of an emerging disease scenario. The dataset is publicly available on GitHub (\url{https://github.com/HAIRLAB/Pre_Surv_COVID_19}).
    \item \textbf{MIMIC-IV Dataset}~\cite{mimic4}: Sourced from the EHRs of the Beth Israel Deaconess Medical Center, it is extensive and widely used in healthcare research, particularly for simulating ICU scenarios. From MIMIC-IV, 17 lab test features and 2 demographic features are extracted. To minimize missing data, we consolidate every consecutive 12-hour segment into a single record for each patient, focusing on the first 48 records.
    % \cw{In problem formulation, features are divided into two kinds: dynamic and static.}
\end{itemize}

We adhere to the benchmark pipeline~\cite{gao2024comprehensive,zhu2023pyehr} for preprocessing both datasets, consistently applying the Last Observation Carried Forward (LOCF)~\cite{wells2013LOCF_Imputation} imputation strategy. This standardized approach facilitates uniformity and comparability in our data analysis. In the context of zero-shot setting, we focus primarily on the test set. For the construction of the test set (same for all models for fair comparison) and training/validation set (specific to ML/DL models that require training), we utilize a stratified shuffled strategy with random selection. Table~\ref{tab:dataset_stats_main_text} presents detailed statistics of the TJH and MIMIC-IV datasets.
% , including the distribution of patients and total visits, along with average visits per patient and length of stay.
Utilizing smaller datasets with around 250 patients and 1000 records is adequate for our zero-shot evaluation, as their representativeness and information-rich records effectively test model generalization in realistic, varied healthcare scenarios. Through the experiments, the MIMIC-IV dataset is processed using secure Azure OpenAI API and human review of the data has been waived.

\begin{table}[htbp]
\centering
\caption{\textit{Statistics of the TJH and MIMIC-IV test datasets.} Avg. visits reports statistics of average visits for each patient. The reported statistics are of the form $Median [Q1, Q3]$.}
\label{tab:dataset_stats_main_text}
\resizebox{\columnwidth}{!}{
\begin{tabular}{lccc}
\toprule
  & Total & Alive & Dead \\
\midrule

\multicolumn{4}{c}{\textit{TJH (test set statistics)}} \\
\midrule
\# Patients & 217 & 117 (53.92 \%) & 100 (46.08 \%)  \\
\# Total Visits   & 986 & 604 (61.26 \%) & 382 (38.74 \%) \\
\# Avg. visits & 5.0 [2.0, 6.0]  & 5.0 [4.0, 6.0] & 3.0 [2.0, 5.0]  \\
Length of stay & 6.0 [2.0, 11.0] & 7.0 [3.0, 12.0] & 4.0[2.0, 8.0] \\

\midrule

\multicolumn{4}{c}{\textit{MIMIC-IV (test set statistics)}} \\
\midrule
\# Patients &  247  & 216 (87.45 \%) & 31 (12.55 \%)  \\
\# Total visits   & 1025 & 895 (87.32 \%) & 130 (12.68 \%) \\
\# Avg. visits & 4.0 [4.0, 4.0]  & 4.0 [4.0, 4.0] & 4.0 [4.0, 4.0]  \\

\bottomrule
\end{tabular}
}
\end{table}

\subsection{Models}
% \yh{TODO}

We have chosen GPT-4~\cite{openai2023gpt4} as our primary model for comparing different prompting strategies due to its dominant performance over other models~\cite{openai2023gpt4}. Unless specifically stated, GPT-4 is employed for experiments in both RQ1 and RQ2. In RQ3, we extend our comparison to include other proprietary LLMs: GPT-3.5~\cite{blog2023ChatGPT}, Gemini Pro~\cite{pichai2023gemini} and open-sources LLM: Llama 2-70B~\cite{touvron2023llama2}. Additionally, in RQ3, we have selected representative ML models: Decision Tree and XGBoost, along with DL models: GRU and AICare~\cite{ma2023aicare}. Notably, AICare is specifically designed for EHR data.

\section{RQ1 (Data-Level) Construction of Prompting Framework for LLMs}\label{sec:rq1}

Given the inherent complexities of EHR data, our study aims to bridge the gap between its three key aspects: 1) structured longitudinal, 2) sparse, and 3) knowledge-infused data, and the unstructured text-based nature LLMs. We conduct our experiments progressively, building upon the best practices concluded from previous experiments. A primary focus of this research is to develop a viable prompt engineering framework that addresses the challenges at the data level, as illustrated in Figure~\ref{fig:prompt_template}.

\subsection{Harnessing Structured Longitudinal EHR}
% \zx{feature-wise(need citing)/visit-wise}

\paragraph{A. Motivation}

\begin{figure}[!ht]
    \centering
    \includegraphics[width=0.95\linewidth]{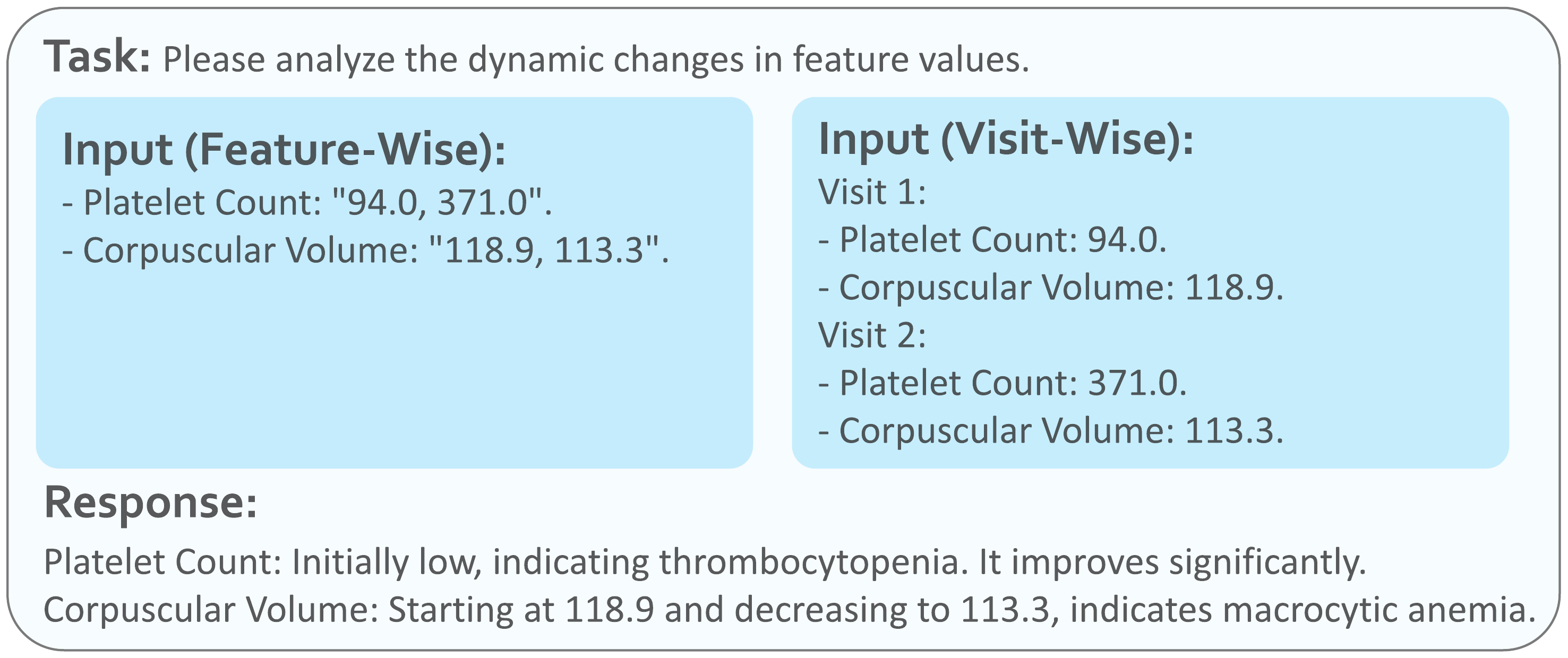}
    \caption{\textit{An example of LLM's analysis with two different input formats representing the longitudinality (Feature-wise \& Visit-wise).}}
    \label{fig:example_input_format_representing_time_series}
\end{figure}

EHRs are inherently structured and longitudinal, capturing patient data over time. This characteristic is at odds with the typical unstructured text-based data formats to which LLMs are accustomed. We aim to investigate how prompts can be effectively structured to represent this time-series data. Structured longitudinal EHR data comprise features over multiple visits for each patient. Naturally, EHR data can be represented from two dimensions: feature-wise and visit-wise, as shown in Figure~\ref{fig:example_input_format_representing_time_series}.

% We explore whether integrating multiple visits into a singular prompt string, as opposed to processing them in batches, can enhance the LLMs' understanding of temporal patient data dynamics. This approach is inspired by the emerging research suggesting LLMs' capabilities as zero-shot time-series learners.

\paragraph{B. Methodology}

We utilize two different input formats to convert the structured longitudinal EHR data into natural language, as described in Figure~\ref{fig:example_input_format_representing_time_series}:

\begin{itemize}[leftmargin=*]
    \item \textbf{Feature-wise:} Present multiple visit data of a patient in one batch. Represent each feature within this data as a string of values, separated by commas.
    \item \textbf{Visit-wise:} Organize multiple visit data of a patient into separate batches, each batch corresponding to one visit.
\end{itemize}

The transformed text is then fed into the LLM using our designed prompt template, which guides the LLM in performing EHR data prediction tasks.

\paragraph{C. Results}

Table~\ref{tab:input_format_perf_longitudinality} reveals that, in the context of in-hospital mortality prediction, the feature-wise format demonstrates superior performance over the visit-wise format in terms of AUROC and AUPRC. We suppose that employing a feature-wise format for inputting EHR data is more effective in enabling LLMs to discern dynamic changes in features, particularly in scenarios characterized by numerous visits and complex variations~\cite{gruver2023LLM_Zeroshot_TS}. Additionally, this method of input requires much fewer tokens with a 44.15\% reduction on TJH and 37.50\% on MIMIC-IV, thereby leading to reduced computational costs.
% \jy{add concrete numbers}
% This approach aligns with the process doctors follow in real clinical scenarios when analyzing a patient's past medical records.

% a test set
%      number   v.w.    f.w.
% tjh   217     834529  466079  +75.05% -44.15%
% mimic 247     202437  126533  +59.99% -37.50%
% total 464     1036966 592612

\begin{table}[!ht]
\caption{\textit{Performance of LLM on in-hospital mortality prediction task with different input formats.} All metrics are multiplied by 100 for readability purposes.}
\label{tab:input_format_perf_longitudinality}
\centering
\resizebox{\columnwidth}{!}{
\begin{tabular}{c|cc|cc}
\toprule
\multirow{2}{*}{Perspective} & \multicolumn{2}{c|}{TJH Mortality} & \multicolumn{2}{c}{MIMIC-IV Mortality} \\
     & AUROC($\uparrow$) & AUPRC($\uparrow$) & AUROC($\uparrow$) & AUPRC($\uparrow$) \\
\midrule
Feature-wise & \textbf{62.29} & \textbf{66.71} & \textbf{60.78} & \textbf{20.95} \\
% & \textbf{89.60} & \textbf{84.18} & \textbf{73.45} & 25.48 \\
Visit-wise   & 57.94 & 58.71 & 57.62 & 19.21 \\          
\bottomrule
\end{tabular}
}
\end{table}

%%%%%%%%%%%%%%%%%%%%%%%%%%%%%%%
%%% 2) Sparsity
%%%%%%%%%%%%%%%%%%%%%%%%%%%%%%%

\subsection{Explore Sparsity in EHR}

\paragraph{A. Motivation}

EHR data often exhibits sparsity, characterized by missing features in the raw data, which poses a challenge for LLMs in understanding these omissions. Traditional ML and DL models typically require imputation before data input, whereas LLMs can directly handle textual inputs. We aim to explore whether the imputation process is necessary for LLMs. 
% and whether LLMs can effectively perceive and interpret missing.

% We intend to compare the outcomes when NaN values are retained versus when they are replaced with Last Observation Carried Forward (LOCF) imputation. This comparison aims to ascertain the most effective method for handling data gaps in EHRs, especially when informing the LLM about imputed values. Understanding how LLMs interpret and process such missing data is crucial for reliable predictions in healthcare.

\paragraph{B. Methodology}

We employ two different methods to address missing values in raw EHR data:

\begin{itemize}[leftmargin=*]
    \item \textbf{Reserving missing values:} Use the term \verb|nan| (``Not a Number'') to signify missing values~\cite{gruver2023LLM_Zeroshot_TS}. We explicitly inform the LLMs about the meaning of `nan' in our prompts.
    \item \textbf{Imputing with LOCF strategy:} Preprocess the raw EHR data by applying the LOCF imputation strategy.
\end{itemize}

Following these preprocessing steps, we transform the EHR data into a natural language format using the feature-wise approach, and then instruct LLMs in performing prediction tasks.

\paragraph{C. Results}

Table~\ref{tab:sparsity_perf} shows our experimental results. Using an imputation strategy does improve the predictive performance of LLMs, though the enhancement observed on MIMIC-IV is not substantial. This indicates that while the imputation process holds some significance for LLMs, reserving NaN in the input EHR data is still adept at interpreting missing values to make predictions. Given that the LOCF strategy involves complex preprocessing, the necessity of applying imputation strategies warrants further exploration.

% and might even be detrimental. We speculate that preserving missing values allows LLMs to analyze the data based solely on available information, thereby avoiding the introduction of potentially misleading data through imputation~\cite{zhu2023leveraging}.

\begin{table}[ht]
\caption{\textit{Performance of LLM on in-hospital mortality prediction task with different methods to } All metrics are multiplied by 100 for readability purposes.}
\label{tab:sparsity_perf}
\centering
\resizebox{\columnwidth}{!}{
\begin{tabular}{c|cc|cc}
\toprule
\multirow{2}{*}{Methods} & \multicolumn{2}{c|}{TJH Mortality} & \multicolumn{2}{c}{MIMIC-IV Mortality} \\
     & AUROC($\uparrow$) & AUPRC($\uparrow$) & AUROC($\uparrow$) & AUPRC($\uparrow$) \\
\midrule
Reserve & 62.29 & 66.71 & 60.78 & 20.95 \\
% & \textbf{89.60} & \textbf{84.18} & \textbf{73.45} & \textbf{25.48} \\
Impute  & \textbf{73.59} & \textbf{69.83} & \textbf{62.72} & \textbf{22.51} \\
\bottomrule
\end{tabular}
}
\end{table}

%%%%%%%%%%%%%%%%%%%%%%%%%%%%%%%
%%% 3) Knowledge-Infused
%%%%%%%%%%%%%%%%%%%%%%%%%%%%%%%

\subsection{Integrating Knowledge-Infused Contexts}

\paragraph{A. Motivation}

\begin{table}[!ht]
    \caption{\textit{Sample EHR data for a single patient visit.}}
    \label{tab:ehr_data_unit_reference_range}
    \centering
        \begin{tabular}{c|c|c|c}
        \toprule
        Test & Value & Reference Range & Units \\ \midrule
        % $\mathrm{Na}$ & 138 & $136-146$ & $\mathrm{mmol} / \mathrm{L}$ \\
        % $\mathrm{K}$ & 4.0 & $3.5-5.0$ & $\mathrm{mmol} / \mathrm{L}$ \\
        $\mathrm{Cl}$ & 105 & 101-109 & $\mathrm{mmol} / \mathrm{L}$ \\
        % $\mathrm{CO_2}$ & 12 & $22-29$ & $\mathrm{mmol} / \mathrm{L}$ \\
        $\mathrm{pH}$ & 7.32 & 7.36-7.41 & / \\
         $\mathrm{P}_\mathrm{CO_2}$ & 40 & $32-45$ & $\mathrm{mmHg}$ \\
        \bottomrule
        \end{tabular}
\end{table}

\begin{figure}[!ht]
\centering
\includegraphics[width=0.95\linewidth]{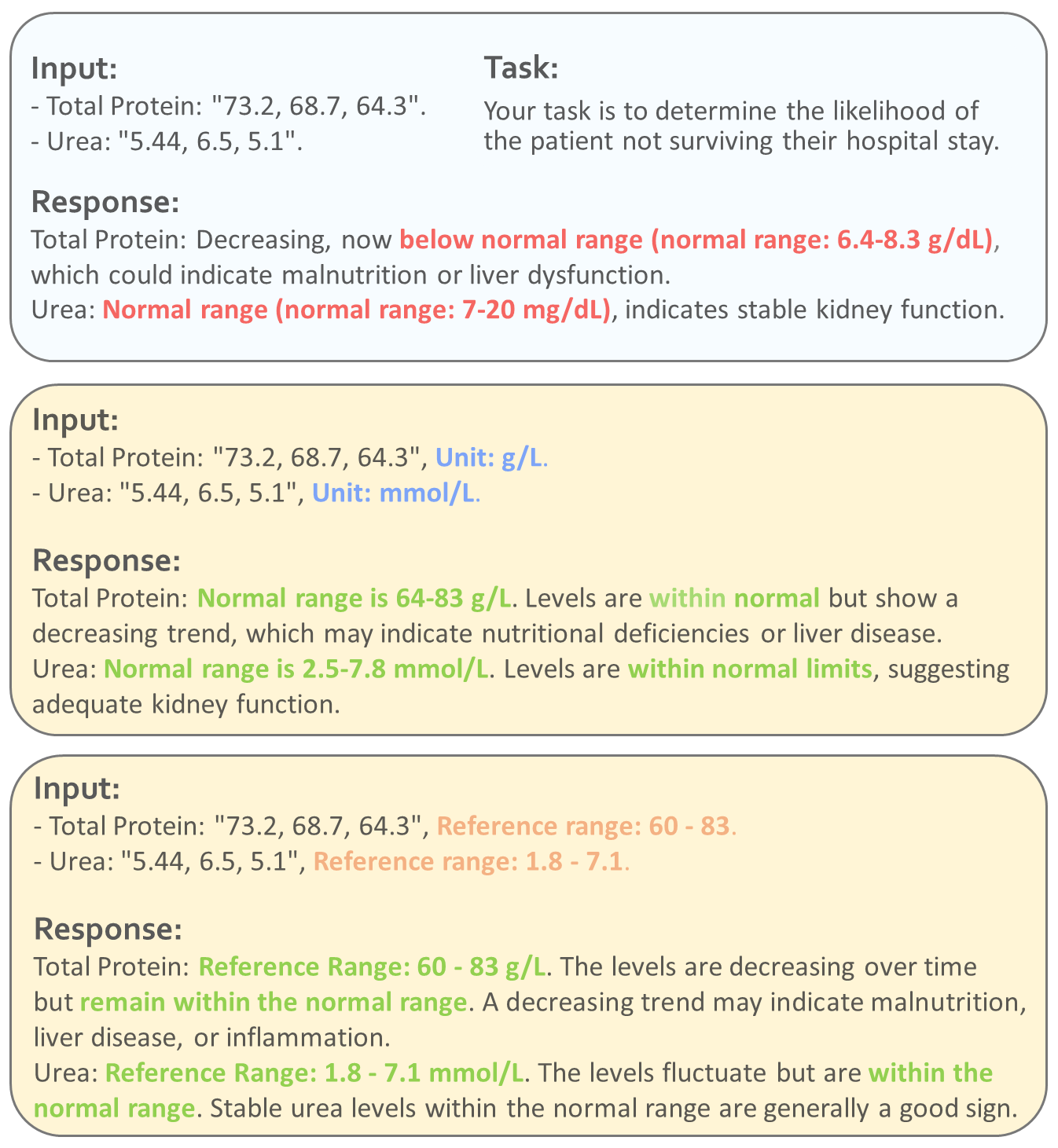}
\caption{\textit{An example of LLM's analysis of a patient's health condition with different context.} 
\textbf{\color{RoseRed}{Red}} stands for incorrect analysis from LLM. \textbf{\color{MyGreen}{Green}} stands for reasonable analysis from LLM.
\textbf{\color{MyBlue}{Blue}} stands for units of features.
\textbf{\color{MyOrange}{Orange}} stands for reference ranges of features.
}
% \jy{text color too light, same fig4}
\label{fig:example_unit_range}
\end{figure}

Table~\ref{tab:ehr_data_unit_reference_range} presents an example of real-world EHR data, highlighting features such as the test name, patient value, reference range, and unit. Intuitively, when EHR data is analyzed without considering them, the vanilla numerical values often become meaningless. As demonstrated in Figure~\ref{fig:example_unit_range}, an LLM's analysis that lacks unit or reference range context can lead to potential misjudgments about a patient's health status.

% Meanwhile, inspired by in-context learning strategy in LLMs, where the model enhances its understanding and response to new prompts using prior examples, we aim to explore whether LLMs can leverage this context by introducing scenarios and examples from past EHR data to improve their predictive performance. This approach also mirrors the clinical decision-making process, where clinicians draw on previous cases and experiences to inform their judgments.
Drawing inspiration from the clinical decision-making process, in which clinicians frequently reference past cases and experiences to inform their judgments on current patients, we apply a similar approach to the in-context learning strategy utilized in prompting LLMs. We aim to investigate the efficacy of LLMs within the healthcare sector, specifically by incorporating scenarios and examples from historical EHR data as contextual anchors, thereby improving their predictive performance.

\paragraph{B. Methodology}

\begin{figure}[!ht]
    \centering
    \includegraphics[width=0.7\linewidth]{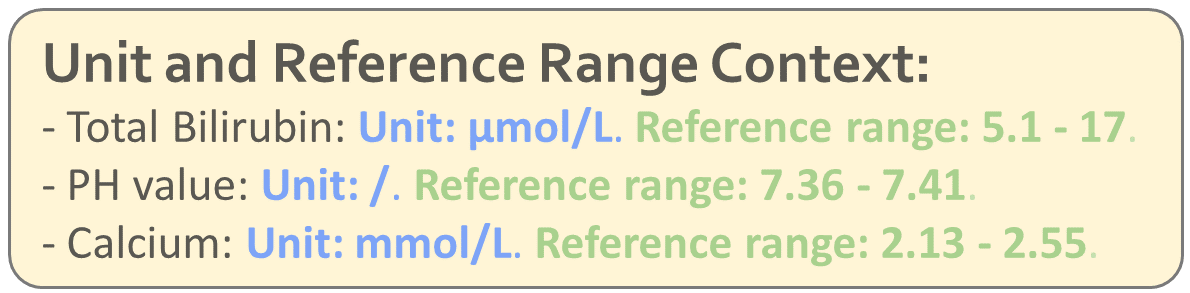}
   \caption{\textit{Prompt of units and reference ranges of sampled features.}}
    \label{fig:unit_range_detail}
\end{figure}

We adopt a streamlined approach to integrate each feature of the dataset into a prompt. As shown in Figure~\ref{fig:unit_range_detail}, this is achieved using a concise template format: \texttt{"- [F]: \{Unit: [U].\}\{Reference range: [R].\}"}, where \texttt{[F]} is used to denote the feature name, while \texttt{[U]} and \texttt{[R]} represent the unit and reference range, respectively. The notation \texttt{\{\}} is employed to indicate that the enclosed content---either the unit or the reference range---is optional.

% \cw{the motivation of in-context learning, why use in-context learning here}
As indicated in the motivation of mirroring the clinical decision-making process, in the in-context learning approach, we also construct the prompt with examples given: \texttt{"Input patient information: [I] RESPONSE: [A]"}, where \texttt{[I]} represents the EHR data formatted in a particular way, and \texttt{[A]} corresponds to the example answer. Here, the sample data is entirely simulated and unrelated to both the training and test sets, thus no issues of data leakage. We aim to identify the optimal number of patient examples for LLMs.

We also use the missing rate metric to evaluate the compliance of LLMs. Missing rate is calculated as $\frac{n_{\text{test}} - n_{\text{decoded}}}{n_{\text{test}}} \times 100\%$, where $n_{\text{test}}$ represents the total number of instances in the test set, and $n_{\text{decoded}}$ denotes the count of instances for which the predicted value was successfully decoded. A lower Missing Rate signifies enhanced performance, reflecting the model's ability to accurately interpret and decode predictions~\cite{xue2023promptcast}.

\paragraph{C. Results}

\begin{table}[!ht]
    \caption{\textit{Performance of LLM on in-hospital mortality prediction task on TJH and MIMIC-IV datasets with different contexts.} \textbf{Bold} indicates the top performance. ``None'' denotes no context. ``M.R.'' denotes ``Missing Rate''. All metrics are multiplied by 100 for readability purposes.}
    \label{tab:unit_range_performance}
    \centering
\begin{tabular}{c|cc|cc}
\toprule
\multirow{2}{*}{Context} & \multicolumn{2}{c|}{TJH Mortality}   & \multicolumn{2}{c}{MIMIC-IV Mortality} \\
       & AUROC($\uparrow$) & M.R.($\downarrow$) & AUROC($\uparrow$) & M.R.($\downarrow$) \\ 
\midrule
None    & 73.59  & 69.83       & 62.72 & 76.11 \\
+Unit   & 87.79  & 20.74       & 65.25 & 50.61 \\
+Range  & 83.57  & \textbf{3.23}        & 68.33 & 9.72  \\
+Both & \textbf{87.83} & 5.07 & \textbf{71.01} & \textbf{9.31}   \\
\bottomrule
\end{tabular}
\end{table}

% \begin{figure}[!ht]
% \centering
% \begin{subfigure}{0.48\linewidth}
%     \centering
%     \includegraphics[width=\linewidth]{figures/unit_range_roc_metric.pdf} % first figure itself
%     \caption{AUROC Metric}
%     \label{fig:unit_range_roc_metric}
% \end{subfigure}
% \begin{subfigure}{0.48\linewidth}
%     \centering
%     \includegraphics[width=\linewidth]{figures/unit_range_missing_rate_metric.pdf}
%     \caption{Missing Rate Metric}
%     \label{fig:unit_range_missing_rate_metric}
% \end{subfigure}
% \caption{\textit{Performance of LLM on in-hospital mortality prediction task on TJH and MIMIC-IV datasets with different contexts.} \textbf{\color{NavyBlue}{Blue}} stands for metric without any context. \textbf{\color{LightBlue}{LightBlue}} stands for metric with unit.
% \textbf{\color{Orange}{Orange}} stands for metric with reference range.
% \textbf{\color{Red}{Red}} stands for metric with unit and reference range.}
% \label{fig:unit_range_perf}
% \end{figure}

% and Figure~\ref{fig:unit_range_perf} \cw{Delete Figure 5}

Table~\ref{tab:unit_range_performance} shows the results of LLM on the mortality prediction task with different contexts, comparing the role of adding units and reference ranges. Adding both of them achieves the best performance with a relative improvement of 19.35\% and 13.22\% in terms of AUROC on both datasets, underscoring their necessities. Notably, when no units and reference ranges are added, the missing rate of LLM predictions is exceptionally high, approaching or exceeding 70\%. And, as units and reference ranges are provided, they help the LLM better comprehend the content of EHR data, thus decreasing the missing rate of its predictions to less than 10\%.

\begin{table}[!ht]
    \caption{\textit{Performance of LLM on in-hospital mortality prediction task on TJH and MIMIC-IV datasets with different numbers of examples.} ``\# Ex.'' means ``Number of examples''. \textbf{Bold} indicates the top performance. All metrics are multiplied by 100 for readability purposes.}
    \label{tab:icl_performance}
    \centering
\resizebox{\columnwidth}{!}{
\begin{tabular}{c|cc|cc}
\toprule
\multirow{2}{*}{\# Ex.} & \multicolumn{2}{c|}{TJH Mortality}   & \multicolumn{2}{c}{MIMIC-IV Mortality} \\
    & AUROC($\uparrow$) & AUPRC($\uparrow$) & AUROC($\uparrow$) & AUPRC($\uparrow$) \\ 
\midrule
0   & 87.83 & 80.45  & 71.01 & 22.29 \\
1   & 91.81 & 84.59  & \textbf{73.96} & \textbf{31.65} \\
2   & \textbf{93.55} & \textbf{88.64} & 71.55 & 24.77 \\
3   & 88.60 & 82.94  & 70.70 & 23.95 \\
\bottomrule
\end{tabular}
}
\end{table}

Table~\ref{tab:icl_performance} shows the performance of the LLM on both datasets with different numbers of examples provided in the context. We find that the inclusion of one or two examples significantly improves the LLM's performance. In the setting of 2 examples on TJH, it reaches 0.9355 on AUROC and 0.8864 on AUPRC, with a relative improvement of 6.51\% on AUROC and 10.18\% on AUPRC compared to no examples. In the setting of 1 example on MIMIC-IV, it also improves by 4.15\% on AUROC and 41.99\% on AUPRC. However, the addition of 3 examples results in a performance decline on both datasets. This issue may arise from the LLM's catastrophic forgetting of the excessively long context of the prompt~\cite{luo2023empirical}.

% \subsection{Sensitivity across Different Temporal Contexts} 

% \section{RQ2 (Task-Level) Are LLMs Capable of Diverse Clinical Prediction Tasks with Various Time Spans?}\label{sec:rq2}
\section{RQ2 (Task-Level) Capability in Diverse Clinical Predictions Over Time}\label{sec:rq2}
\paragraph{A. Motivation}

The diverse nature of clinical scenarios necessitates predictive models flexible and robust, capable of adapting to various healthcare needs. Our focus is on evaluating the ability of LLMs to manage a spectrum of prediction tasks. Clinical scenarios range from acute conditions requiring immediate attention to more routine predictive tasks for chronic diseases. Thus, we also aim to assess the sensitivity and efficacy of LLMs across different time spans, providing valuable insights for both short- and long-term patient management.
% \yh{TODO: cue sensitivity}

\paragraph{B. Methodology}

Our methodology focuses primarily on evaluating LLMs across various clinical predictive tasks, with an auxiliary examination of their temporal sensitivity:

\begin{itemize}[leftmargin=*]
    \item \textbf{Evaluation of Diverse Predictive Tasks}: These tasks include in-hospital mortality, 30-day readmission, length-of-stay and multi-task predictions tasks.
    \item \textbf{Temporal Sensitivity Observation}: As a supplementary observation, we will examine the LLMs' performance in predicting in-hospital mortality at different time spans: upon discharge, one month post-discharge, and six months post-discharge.
\end{itemize}

These tasks modify the instruction part of the prompt template, as illustrated in Figure~\ref{fig:prompt_template}, reflecting the task-level assessment.

\paragraph{C. Results}

\begin{table}[htbp]
\caption{\textit{Performance of LLMs on length-of-stay prediction task on TJH.} \textbf{Bold} indicates the best performance. Note that the results contain unknown predictions. For unknown cases, we set the value of 0.}
\label{tab:llm_los_perf}
\centering
% \resizebox{\columnwidth}{!} {
\begin{tabular}{c|cccc}
\toprule
Methods    & MAE($\downarrow$)  & MSE($\downarrow$)   & RMSE($\downarrow$) \\
\midrule
Gemini Pro & 6.03 & 79.77 & 8.93 \\
GPT-3.5    & 5.78 & 70.47 & 8.39 \\
GPT-4      & \textbf{5.53} & \textbf{68.21} & \textbf{8.26} \\
\bottomrule
\end{tabular}
% }
\end{table}

Table~\ref{tab:llm_los_perf} presents the performance of three closed-sourced LLMs on the length-of-stay prediction task on the TJH dataset. Notably, GPT-4 outperforms the other two models. Excluding samples where GPT-4 can not predict, it achieves an MAE of 3.47, surpassing the previous best performance (3.58) reported in COVID-19 Benchmarks~\cite{gao2024comprehensive}.

Table~\ref{tab:mimic_object_level_perf} details the performance of LLMs across various tasks on the MIMIC-IV dataset. The results highlight LLMs' exceptional ability to follow prompts in diverse prediction scenarios. However, it is observed that their ability to perform multiple tasks simultaneously (multi-task setting) can result in a reduction in overall performance.

\begin{table}[htbp]
\caption{\textit{Performance of LLMs on different tasks on MIMIC-IV.} All metrics are multiplied by 100 for readability purposes.}
\label{tab:mimic_object_level_perf}
\centering
\begin{tabular}{c|c|cc}
\toprule
\multicolumn{2}{c|}{Setting} & AUROC($\uparrow$) & AUPRC($\uparrow$) \\
\midrule
\multirow{2}{*}{Single-task} 
& Mortality   & 73.96 & 31.65 \\
& Readmission & 68.61 & 28.23 \\
\midrule
\multirow{2}{*}{Multi-task}   
& Mortality   & 71.69 & 23.16 \\       
& Readmission & 64.07 & 22.61 \\
\bottomrule
\end{tabular}
\vskip -1em
\end{table}

\begin{figure}[htbp]
\centering
\begin{subfigure}{0.48\linewidth}
  \centering
  \includegraphics[width=\linewidth]{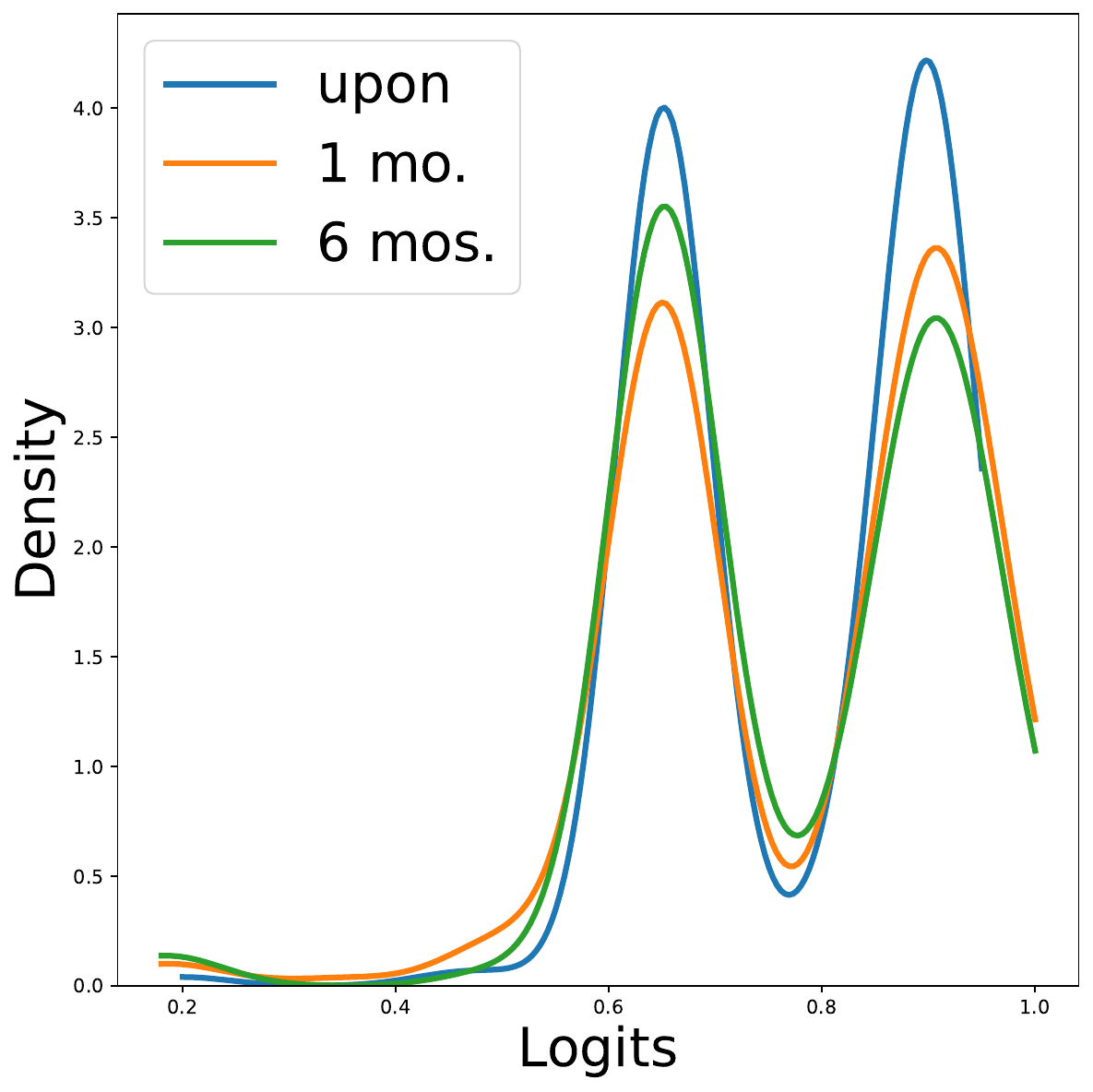}
  \caption{TJH}
  \label{fig:tjh_time_level_distribution}
\end{subfigure}
\begin{subfigure}{0.48\linewidth}
  \centering
  \includegraphics[width=\linewidth]{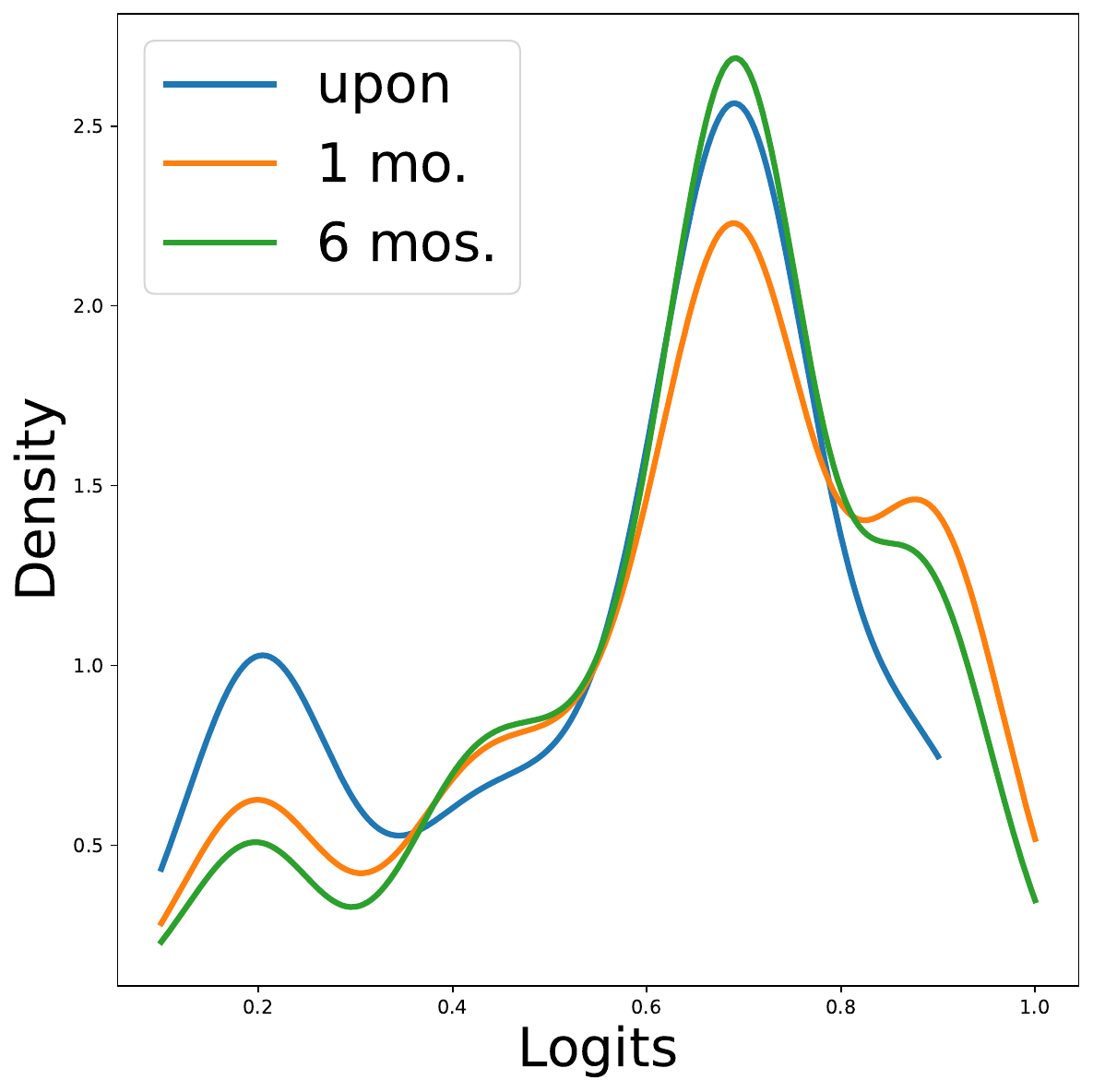}
  \caption{MIMIC-IV}
  \label{fig:mimic_time_level_distribution}
\end{subfigure}
\caption{\textit{Density of prediction logits on TJH and MIMIC-IV datasets across three time spans: upon discharge, 1 month and 6 months post-discharge.}}
\label{fig:time_level_distribution}
\end{figure}

Figure~\ref{fig:time_level_distribution} illustrates the distribution of prediction logits across three different time spans on both datasets. It is worth noting that the prediction logits exhibit similar peaks for all three time spans on both datasets, showing no significant variances. This pattern suggests that LLMs may not exhibit heightened sensitivity to the time dimension on prediction tasks.

% \yh{first illustrate the figure (similar peaks)} 
% \yh{Criticise LLM}

\begin{table*}[htbp]
% \jy{wrong table order?}
\caption{\textit{Performance of selected models on TJH and MIMIC-IV datasets in-hospital mortality prediction task.} ``O.S.'' denotes ``Open Source'' and ``C.S.'' means ``Closed Source''. \textbf{Bold} indicates our elaborated designed prompting framework shows better performance than basic prompts. \textbf{\textcolor{red}{Red}} denotes best performance among all baselines of all settings. All metrics are multiplied by 100 for readability purposes.}
\label{tab:benchmarking_performance_overall}
\centering
\resizebox{\linewidth}{!}{
\begin{tabular}{c|cc|c|ccc|ccc}
\toprule
 \multicolumn{3}{c|}{\multirow{2}{*}{Methods}} & \multirow{2}{*}{Shots} & \multicolumn{3}{c|}{TJH Mortality} & \multicolumn{3}{c}{MIMIC-IV Mortality} \\
 \multicolumn{3}{c|}{} & & AUROC($\uparrow$) & AUPRC($\uparrow$) & min(+P, Se)($\uparrow$) & AUROC($\uparrow$) & AUPRC($\uparrow$) & min(+P, Se)($\uparrow$) \\ \midrule
\multirow{3}{*}{ML/DL} 
 & \multicolumn{2}{c|}{Decision Tree} & 10 & 65.78 & 56.18 & 59.37 & 57.27 & 14.59 & 16.07 \\
 & \multicolumn{2}{c|}{XGBoost} & 10 & 85.01 & 80.65 & 76.00 & 53.41 & 14.40 & 20.41 \\
 & \multicolumn{2}{c|}{GRU} & 10 & 70.47 & 66.24 & 60.00 & 49.81 & 16.35 & 16.42 \\
 & \multicolumn{2}{c|}{AICare} & 10 & 77.75 & 73.30 & 68.31 & 43.49 & 13.62 & 13.26 \\
\midrule
\multirow{1}{*}{O.S.} & \multicolumn{2}{c|}{Llama 2-70B} & 0 & 52.66 & 50.53 & 50.35 & 60.39 & 20.53 & 24.49 \\ \midrule
\multirow{6}{*}{C.S.} 
& \multirow{2}{*}{Gemini Pro}
& base & 0 & 61.05 & 57.02 & 54.35 & \multicolumn{3}{c}{\multirow{2}{*}{Not tested due to privacy issues}} \\
& & ours & 0 & \textbf{78.89} & \textbf{71.94} & \textbf{70.19} & \multicolumn{3}{c}{} \\
% \midrule
\cmidrule{2-10} 
& \multirow{2}{*}{GPT-3.5}
& base & 0 & 60.38 & 57.32 & 50.34 & 54.85 & 15.23 & 16.13 \\
& 
& ours & 0 & \textbf{76.02} & \textbf{80.10} & \textbf{68.97} & \textbf{70.85} & \textbf{26.92} & \textbf{26.39} \\
% \midrule
\cmidrule{2-10} 
& \multirow{2}{*}{GPT-4} 
& base & 0 & 62.29 & 66.71 & 49.08 & 60.78 & 20.95 & 32.26 \\
& & ours & 0 & \textcolor{red}{\textbf{93.55}} & \textcolor{red}{\textbf{88.64}} & \textcolor{red}{\textbf{87.00}} & \textcolor{red}{\textbf{73.96}} & \textcolor{red}{\textbf{31.65}} & \textcolor{red}{\textbf{39.39}} \\ 
\bottomrule
\end{tabular}
}
\end{table*}

\section{RQ3 (Model-Level) Benchmarking Zero-Shot Performance of LLMs}\label{sec:rq3}
% \section{RQ3 (Model-Level) What Are the Zero-shot Performance Differences of Various LLMs in Handling EHR Data, and Can They Outperform Traditional Machine Learning Models in Few-shot Settings?}\label{sec:rq3}

% \yh{TODO: 1) Briefly describe ML/DL/LLM models in one line. 2) setting difference 10 shots (5 positive, 5 negative case) and our LLM's 0 shots (and their setting, follow best practice on previous RQ) 3) describe performance 4) conclusion 5) analyze GPT-4 beats all}

% baseline models?
% \begin{itemize}
%     \item 
% \end{itemize}

\paragraph{A. Motivation}

Traditional ML and DL models have established baselines in the realm of EHR data analysis, typically requiring substantial datasets for training. However, in many clinical situations, especially with emerging diseases or rare conditions, the availability of extensive data is not a given. In this context, our research is driven by the need to ascertain whether LLMs, particularly known for their zero-shot learning capabilities, can outperform these traditional ML/DL models in scenarios with restricted data availability. This exploration is pivotal to our research as it directly addresses the challenge of `cold start' in clinical prediction tasks involving EHR data.

\paragraph{B. Methodology}

In the case of traditional ML and DL models, training occurs in a 10-shot setting. We randomly select 5 positive and 5 negative cases to form the training set, ensuring that none of these patients are included in the test set. Subsequently, we assess their performance using the same test set that is applied to LLMs. For LLMs in a zero-shot setting, we adhere to the best practices established in prior research questions. We guide LLMs to execute prediction tasks by transforming EHR data into a natural language format on a feature-wise basis, incorporating units, reference ranges, and an example of a simulated input-output pair in the prompt to provide clinical context. This approach forms the basis of our prompting framework as depicted in Figure~\ref{fig:prompt_template}.

\paragraph{C. Results}

Table~\ref{tab:benchmarking_performance_overall} displays the performance of all baseline models, which include traditional ML and DL models, as well as open-sourced and closed-sourced LLMs, in the in-hospital mortality prediction task on the TJH and MIMIC-IV datasets. Within our tailored prompting framework, GPT-4 outperforms other models across all metrics on both datasets. The average relative improvements in the AUROC metric are 35.10\% and 25.43\%, and for the AUPRC metric, they are 32.93\% and 63.92\% on the TJH and MIMIC-IV datasets, respectively, confirming the efficacy of our framework. Remarkably, GPT-4, operating as a zero-shot model, shows superior performance compared to traditional ML models. Other LLMs also demonstrate high performance on the MIMIC-IV mortality task and are competitive in the TJH mortality task. However, GPT-4 stands out significantly, even matching the performance achieved with the full training set in previous benchmarks on the TJH dataset~\cite{gao2024comprehensive}, indicating a substantial gap in effectiveness. Therefore, we posit that while other LLMs have room for improvement, GPT-4, as a leading model, shows potential for clinical application. In summary, this experiment underscores the capability of LLMs in scenarios with extremely limited records.

% GPT-4 surpasses all baseline methods on both datasets from all metrics. 
% 传统的机器学习模型，在样本量不足的情况下，无法学习到足够的知识以支持其准确预测；相比之下，开源大语言模型在zero-shot的setting下，仍然可以通过高效的prompt strategy达到优秀的性能表现。这证明了LLM在EHR 任务上同样具有良好的zero-shot能力，这具有十分重要的意义。

\section{Discussions}\label{sec:discussions} 

% \yh{If out of space, put them into Appendix}

\paragraph{Limitations} In our study, the model primarily outputs a single logit value for each prediction in all experimental settings. While efficient, it does limit the depth of insight into the underlying reasoning behind these predictions, an aspect that merits consideration in the context of clinical decision-making. Furthermore, incorporating a range of recent LLMs with varying parameter sizes could potentially enrich our analysis, offering a more nuanced understanding of these models' capabilities.

\paragraph{Future Work} Our future work includes considering the application of chain-of-thought reasoning, improved strategies for managing missing values, and the implementation of a self-consistency method to augment zero-shot clinical predictions using EHR data. Additionally, investigating the uncertainties inherent in the outputs of LLMs and input EHR data is of substantial interest.

Detailed implementations and the full prompt template can be found in the Appendix.
% \cw{section 7 is short and can be merged into section 8}
\section{Conclusions}\label{sec:conclusions}

% \yh{TOFIX}

We have successfully integrated Large Language Models (LLMs) with structured longitudinal EHR data using our prompting framework, achieving superior performance in zero-shot clinical predictions. \textbf{In RQ1 at the data level}, we demonstrate that combining clinical contexts with in-context learning strategies, mirroring clinicians' diagnostic procedures, effectively mitigates numerical understanding issues. \textbf{In RQ2 at the task level}, our findings indicate that LLMs can identify and respond to diverse task requirements in prompts. However, they exhibit limited sensitivity to varying time spans, as evidenced in ICU and chronic disease predictions. \textbf{In RQ3 at the model level}, benchmarking different LLMs shows that GPT-4's zero-shot mortality predictions outperform those of machine learning models trained with limited data, highlighting LLMs' potential in enhancing clinical predictions in scenarios lacking labeled data.

\section*{Ethical Statement}

This study, involving the analysis of Electronic Health Records (EHR) using the MIMIC dataset, is committed to upholding high ethical standards. The MIMIC dataset is a de-identified dataset, ensuring patient confidentiality and privacy. It is available through a data use agreement, underscoring our commitment to responsible data handling and usage. The MIMIC dataset is processed using secure Azure OpenAI API and human review of the data has been waived.

\section*{Acknowledgments}

This work is supported by the National Key R\&D Program of China (No. 2022ZD0116401).

\clearpage
\newpage

\bibliographystyle{named}
\bibliography{ref}

\clearpage
\newpage
\appendix

\onecolumn

\section{Appendix}

\subsection{Generation of Units and Reference Ranges Details}

Initially, we employ a simple prompt, instructing the LLM (GPT-4) to generate units and reference ranges for all features in both datasets. The LLM is able to infer over 80\% of the features, demonstrating its capability to identify and associate its inherent clinical-context knowledge with the features in most cases. Furthermore, upon a detailed review of the LLM's outputs, we find that some of the units generated by the LLM differ from those used in the datasets. Consequently, we refer to authoritative medical literature for minor adjustments to the LLM-generated reference ranges, thereby ensuring higher accuracy and relevance. In instances where medical literature does not provide the necessary reference ranges, we adopt an alternative statistical approach by defining the appropriate reference ranges based on the 25th to 75th percentile values of the feature statistics within the datasets. The exported units and reference ranges of features are shown in Table~\ref{tab:units_and_range_in_mimic} and~\ref{tab:units_and_range_in_tjh}.

\begin{table}[htbp]
    \centering
    \caption{\textit{Units and reference ranges of features in MIMIC-IV dataset.} ``GCS'' means ``Glascow coma scale''. ``bpm'' means ``breaths per minute''.}
    \label{tab:units_and_range_in_mimic}
\begin{tabular}{c|c|c}
\toprule
Feature & Unit & Reference Range \\
\midrule
Capillary refill rate           	&	/               	&	/	\\
GCS eye opening  	&	/               	&	/	\\
GCS motor response	&	/               	&	/	\\
GCS total        	&	/               	&	/	\\
GCS verbal response	&	/               	&	/	\\
Diastolic blood pressure        	&	mmHg            	&	less than 80	\\
Fraction inspired oxygen        	&	/               	&	more than 0.21	\\
Glucose                         	&	mg/dL           	&	70 - 100	\\
Heart Rate                      	&	bpm             	&	60 - 100	\\
Height                          	&	cm              	&	/	\\
Mean blood pressure             	&	mmHg            	&	less than 100	\\
Oxygen saturation               	&	\%              	&	95 - 100	\\
Respiratory rate                	&	bpm	                &	15 - 18	\\
Systolic blood pressure         	&	mmHg            	&	less than 120	\\
Temperature                     	&   $^\circ$C 	
&	36.1 - 37.2	\\
Weight                          	&	kg              	&	/	\\
pH                              	&	/               	&	7.35 - 7.45	\\
\bottomrule 
\end{tabular}
\end{table}

\begin{longtable}{c|c|c}
\caption{\textit{Units and reference ranges of features in TJH dataset.}} 
\label{tab:units_and_range_in_tjh} \\
\toprule
Feature & Unit & Reference Range \\
\midrule
\endfirsthead

\multicolumn{3}{c}%
{\tablename\ \thetable\ -- \textit{Continued from previous page}} \\
\midrule
Feature & Unit & Reference Range \\
\midrule
\endhead

\midrule \multicolumn{3}{r}{\textit{Continued on next page}} \\
\endfoot

\bottomrule
\endlastfoot

% Your table data starts here
Hypersensitive cardiac troponinI	&	ng/L            	&	less than 14	\\
hemoglobin                      	&	g/L             	&	140 - 180 for men, 120 - 160 for women	\\
Serum chloride                  	&	mmol/L          	&	96 - 106	\\
Prothrombin time                	&	seconds         	&	13.1 - 14.125	\\
procalcitonin                   	&	ng/mL           	&	less than 0.05	\\
eosinophils(\%)                  	&	\%               	&	1 - 6	\\
Interleukin 2 receptor          	&	pg/mL           	&	less than 625	\\
Alkaline phosphatase            	&	IU/L            	&	44 - 147	\\
albumin                         	&	g/dL            	&	3.5- 5.5	\\
basophil(\%)                     	&	\%               	&	0.5 - 1	\\
Interleukin 10                  	&	pg/mL           	&	less than 9.8	\\
Total bilirubin                 	&	$\mu$mol/L          &	5.1 - 17	\\
Platelet count                  	&	$\times$ 10$^9$/L   &	150 - 450	\\
monocytes(\%)                    	&	\%               	&	2 - 10	\\
antithrombin                    	&	\%               	&	80 - 120	\\
Interleukin 8                   	&	pg/mL           	&	less than 62	\\
indirect bilirubin              	&	$\mu$mol/L          &	3.4 - 12.0	\\
Red blood cell distribution width	&	\%               	&	11.5 - 14.5 for men, 12.2 - 16.1 for women	\\
neutrophils(\%)                  	&	\%               	&	45 - 70	\\
total protein                   	&	g/L             	&	60 - 83	\\
Quantification of Treponema pallidum antibodies	&	/       &	less than 1.0	\\
Prothrombin activity            	&	\%               	&	70 - 130	\\
HBsAg                           	&	IU/mL           	&	0.0 - 0.01	\\
mean corpuscular volume         	&	fL              	&	80 - 100	\\
hematocrit                      	&	\%               	&	40 - 54 for men, 36 - 48 for women	\\
White blood cell count          	&	$times$ 10$^9$/L    &	4.5 - 11.0	\\
Tumor necrosis factorα          	&	pg/mL           	&	less than 8.1	\\
mean corpuscular hemoglobin concentration	&	g/L         &	320 - 360	\\
fibrinogen                      	&	g/L             	&	2 - 4	\\
Interleukin 1β                  	&	pg/mL           	&	less than 6.5	\\
Urea                            	&	mmol/L          	&	1.8 - 7.1	\\
lymphocyte count                	&	$\times$ 10$^9$/L   &	1.0 - 4.8	\\
PH value                        	&	/               	&	7.35 - 7.45	\\
Red blood cell count            	&	$\times$ 10$^12$/L  &	4.5 - 5.5 for men, 4.0 - 5.0 for women	\\
Eosinophil count                	&	$\times$ 10$^9$/L   &	0.02 - 0.5	\\
Corrected calcium               	&	mmol/L          	&	2.12 - 2.57	\\
Serum potassium                 	&	mmol/L          	&	3.5 - 5.0	\\
glucose                         	&	mmol/L          	&	3.9 - 5.6	\\
neutrophils count               	&	$\times$ 10$^9$/L   &	2.0 - 8.0	\\
Direct bilirubin                	&	$\mu$mol/L          &	1.7 - 5.1	\\
Mean platelet volume            	&	fL              	&	7.4 - 11.4	\\
ferritin                        	&	ng/mL           	&	24 - 336 for men, 11 - 307 for women	\\
RBC distribution width SD       	&	fL              	&	40 - 55	\\
Thrombin time                   	&	seconds         	&	12 - 19	\\
(\%)lymphocyte                   	&	\%               	&	20 - 40	\\
HCV antibody quantification     	&	IU/mL           	&	0.04 - 0.08	\\
DD dimer                        	&	mg/L            	&	0 - 0.5	\\
Total cholesterol               	&	mmol/L          	&	less than 5.17	\\
aspartate aminotransferase      	&	U/L             	&	8 - 33	\\
Uric acid                       	&	$\mu$mol/L          &	240 - 510 for men, 160 - 430 for women	\\
HCO3                            	&	mmol/L          	&	22 - 29	\\
calcium                         	&	mmol/L          	&	2.13 - 2.55	\\
Aminoterminal brain natriuretic peptide precursor(NTproBNP)	&	pg/mL           	&	0 - 125	\\
Lactate dehydrogenase           	&	U/L             	&	140 - 280	\\
platelet large cell ratio       	&	\%               	&	15 - 35	\\
Interleukin 6                   	&	pg/mL           	&	0 - 7	\\
Fibrin degradation products     	&	μg/mL           	&	0 - 10	\\
monocytes count                 	&	$\times$ 10$^9$/L        	&	0.32 - 0.58	\\
PLT distribution width          	&	fL              	&	9.2 - 16.7	\\
globulin                        	&	g/L             	&	23 - 35	  \\
$\gamma$glutamyl transpeptidase        	&	U/L             	&	7 - 47 for men, 5 - 25 for women	\\
International standard ratio    	&	ratio           	&	0.8 - 1.2	\\
basophil count(\#)               	&	$\times$ 10$^9$/L   &	0.01 - 0.02	\\
mean corpuscular hemoglobin     	&	pg              	&	27 - 31	\\
Activation of partial thromboplastin time	&	seconds     &	22 - 35	\\
High sensitivity Creactive protein	&	mg/L            	&	3 - 10	\\
HIV antibody quantification     	&	IU/mL           	&	0.08 - 0.11	\\
serum sodium                    	&	mmol/L          	&	135 - 145	\\
thrombocytocrit                 	&	\%               	&	0.22 - 0.24	\\
ESR                             	&	mm/hr           	&	less than 15  for men, less than 20 for women	\\
glutamicpyruvic transaminase    	&	U/L             	&	0 - 35	\\
eGFR                            	&	mL/min/1        	&	more than 90	\\
creatinine                      	&	$\mu$mol/L          &	61.9 - 114.9 for men, 53 - 97.2 for women	\\
% Repeat the pattern for as many rows as you need
\end{longtable}

\subsection{Generation of Examples for In-context Learning Details}

Our approach centers on supplying LLMs with standardized input-output examples, where the data content is entirely simulated. The primary objective is to identify the optimal number of examples needed for the LLMs to deliver stable and effective predictions. Initially, we secure a sample set through random sampling, ensuring it excludes any patients present in the test set. We then divide this sample set based on patient mortality outcomes. For both the surviving and deceased patient groups, we compute the average and variance of feature values derived from all their visits. Utilizing the Gaussian random function, we generate a sequence of random feature values to represent sample patients. Concurrently, we assign each sample patient a random floating-point number to indicate their mortality risk. This number ranges from 0 to 0.5 for survivors and from 0.5 to 1 for deceased patients (applicable for mortality prediction tasks, and similarly for other tasks). We then feed these values to LLMs using our designed template.

\subsection{Task Description Details}

\begin{figure}[htbp]
    \centering
    \includegraphics[width=0.5\linewidth]{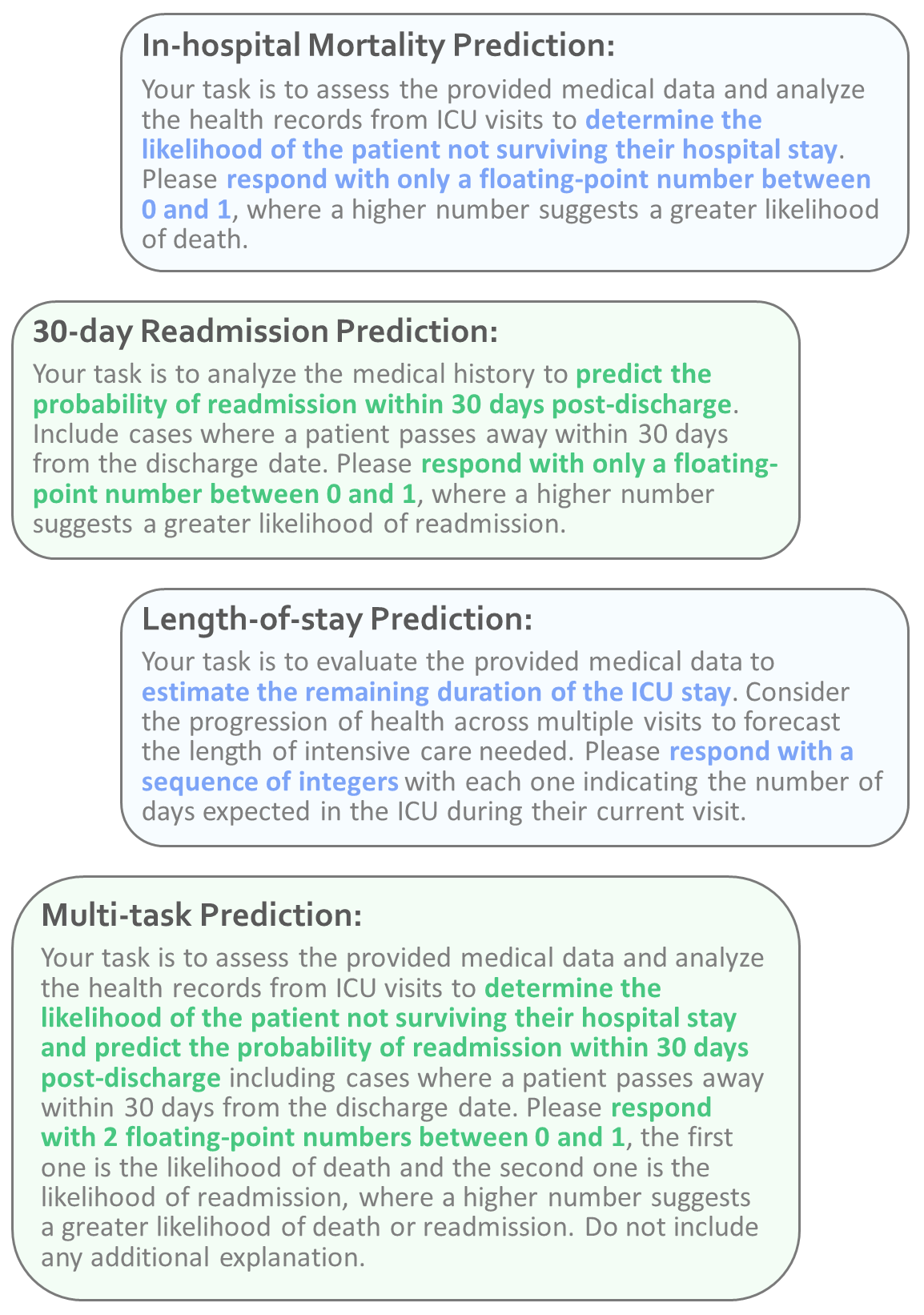}
    \caption{\textit{Prompt of task description and response format on different tasks.}}
    \label{fig:tasks_prompt}
\end{figure}

We design task descriptions for various predictive tasks, including in-hospital mortality, 30-day readmission, length-of-stay and multi-task prediction tasks, as shown in Figure~\ref{fig:tasks_prompt}. These task descriptions modify the instruction part of our designed prompt template.
% \yh{TODO: description of figures}

\subsection{Implementation Details}

For machine learning or deep learning model baselines illustrated in benchmarked performance table, we train these models on a server equipped with Nvidia RTX 3090 GPU and 64GB RAM. The software environment is CUDA 12.2, Python 3.11, PyTorch 2.0.1, PyTorch Lightning 2.0.5. We use AdamW optimizer. All models are trained via 50 epochs over patient samples on the training set, and the early stop strategy monitored by AUPRC with 10 epochs is applied.

% \yh{TODO: OpenAI GPT version, approximate costs}
Our experiments are carried out over a period spanning from December 19th, 2023, to January 17th, 2024. We employ GPT-4 Turbo (GPT-4-1106-Preview), GPT-3.5 Turbo (GPT-3.5-Turbo-1106), and Gemini Pro as our primary baseline models for this study. The overall estimated expenditure for conducting these experiments amounts to \$272.14.

% \subsection{CoT Strategy}
\subsection{Apply Chain-of-Thought (CoT) Prompting}
% \yh{Reasong, in addition to the simple logits. Full version in appendix (prompt of CoT step-by-step thinking), the reasoning results are consistent with medical literature}
% Chain-of-Thought Prompting~\cite{Wei2022CoT}, involves guiding LLMs to process information step-by-step towards a solution. This technique has proven effective in enhancing LLM performance across various tasks (i.e. mathematical problem solving). We incorporated the CoT strategy into our EHR tasks as well. However, in the context of in-hospital mortality prediction on the MIMIC-IV dataset, the optimal setting from previous research questions achieved an AUROC of 0.7396, whereas the implementation of the CoT strategy resulted in a reduced AUROC of 0.7149. While the reasoning provided by the LLM aligns with medical literature, it does not translate into more accurate predictions. A potential explanation is that LLMs, when employing CoT, tend to focus on specific patterns with abnormal values, overlooking broader feature relationships that are crucial in EHR data prediction tasks. We list the prompt template in CoT strategy below:

Chain-of-Thought Prompting (CoT)~\cite{Wei2022CoT} guides LLMs to process information step-by-step toward a solution. This technique effectively enhances LLM performance in various tasks, such as mathematical problem-solving. We also incorporate the CoT strategy in our EHR tasks. However, in the context of in-hospital mortality prediction using the MIMIC-IV dataset, the optimal setting from previous research achieved an AUROC of 0.7396. In contrast, implementing the CoT strategy resulted in a lower AUROC of 0.7149. Although the reasoning provided by the LLM aligns with medical literature, it does not yield more accurate predictions. A potential explanation is that LLMs, when using CoT, tend to focus on specific patterns with abnormal values, overlooking broader feature relationships crucial in EHR data prediction tasks. Below is the prompt template used in the CoT strategy:

% \yh{Possible reason: When reasoning, LLM only output certain selected patterns, while neglecting the relationships with more features.}

% \begin{lstlisting}
% \end{lstlisting}

\begin{tcolorbox}[title=Complete Prompt Template for Optimal Settings in the TJH Dataset,
    colback=white,
    colframe=black,
    colbacktitle=white,
    coltitle=black,
    fonttitle=\bfseries,
    breakable,
    enhanced]
% \begin{verbatim}
% \begin{BVerbatim}[fontsize=\small]
\begin{Verbatim}[fontsize=\small]
Please follow the Chain-of-Thought Analysis Process:

1. Analyze the data step by step, For example:
   - Blood pressure shows a slight downward trend, indicating...
   - Heart rate is stable, suggesting...
   - Lab results indicate [specific condition or lack thereof]...
   - The patient underwent [specific intervention], which could mean...

2. Make Intermediate Conclusions:
   - Draw intermediate conclusions from each piece of data. For example:
     - If a patient's blood pressure is consistently low, it might indicate poor 
       cardiovascular function.
     - The patient's cardiovascular function is [conclusion].
     - [Other intermediate conclusions based on data].

3. Aggregate the Findings:
   - After analyzing each piece of data, aggregate these findings to form a comprehensive 
     view of the patient's condition.
   - Summarize key points from the initial analysis and intermediate conclusions.

Aggregated Findings:
- Considering the patient's vital signs and lab results, the overall health status is...

4. Final Assessment:
   - Conclude with an assessment of the patient's likelihood of not surviving their 
     hospital stay.
   - Provide a floating-point number between 0 and 1, where a higher number suggests a 
     greater likelihood of death.
   - If the data is insufficient or ambiguous, conclude with "I do not know."

[0.XX] or "I do not know."

Example Chain-of-Thought Analysis:

1. Analyze the data step by step:
   - Blood pressure shows a slight downward trend, which might indicate a gradual decline 
     in cardiovascular stability.
   - Heart rate is stable, which is a good sign, suggesting no immediate cardiac distress.
   - Elevated white blood cell count could indicate an infection or an 
     inflammatory process in the body.
   - Low potassium levels might affect heart rhythm and overall muscle function.

2. Make Intermediate Conclusions:
   - The decreasing blood pressure could be a sign of worsening heart function or 
     infection-related hypotension.
   - Stable heart rate is reassuring but does not completely rule out underlying issues.
   - Possible infection, considering the elevated white blood cell count.
   - Potassium levels need to be corrected to prevent cardiac complications.

3. Aggregate the Findings:
   - The patient is possibly facing a cardiovascular challenge, compounded by an infection 
     and electrolyte imbalance.

Aggregated Findings:
- Considering the downward trend in blood pressure, stable heart rate, signs of infection, 
  and electrolyte imbalance, the patient's overall health status seems to be moderately 
  compromised.

4. Final Assessment:
0.65
\end{Verbatim}
% \end{BVerbatim}
% \end{verbatim}
\end{tcolorbox}

\subsection{Comprehensive Prompt Templates for Optimal Settings Across Both Datasets}
% \zx{Both Datasets and Best Settings}
% \yh{TODO: description}

We have listed two comprehensive prompt templates that yielded the best performance on the TJH and MIMIC-IV datasets' mortality prediction task in our previous research:

\begin{itemize}[leftmargin=*]
    \item \textbf{TJH Dataset Mortality Prediction Task:}
% \begin{lstlisting}
% \end{lstlisting}
% \begin{tcolorbox}[title=Complete Prompt Template for Optimal Settings in the TJH Dataset,
%     colback=white,
%     colframe=black,
%     colbacktitle=white,
%     coltitle=black,
%     fonttitle=\bfseries,
%     breakable,
%     enhanced]
% % \begin{minted}
% % [
% % frame=lines,
% % framesep=2mm,
% % baselinestretch=1.2,
% % fontsize=\footnotesize,
% % ]
% % {text}
% \begin{lstlisting}

\begin{tcolorbox}[title=Complete Prompt Template for Optimal Settings in the TJH Dataset,
    colback=white,
    colframe=black,
    colbacktitle=white,
    coltitle=black,
    fonttitle=\bfseries,
    breakable,
    enhanced]
% \begin{verbatim}
% \begin{BVerbatim}[fontsize=\small]
\begin{Verbatim}[fontsize=\small]
You are an experienced doctor in Intensive Care Unit (ICU) treatment.

I will provide you with medical information from multiple Intensive Care Unit (ICU) 
visits of a patient, each characterized by a fixed number of features.

Present multiple visit data of a patient in one batch. Represent each feature within 
this data as a string of values, separated by commas.

Your task is to assess the provided medical data and analyze the health records from 
ICU visits to determine the likelihood of the patient not surviving their hospital 
stay. 
Please respond with only a floating-point number between 0 and 1, where a higher 
number suggests a greater likelihood of death.

In situations where the data does not allow for a reasonable conclusion, respond with 
the phrase `I do not know` without any additional explanation.

- Hypersensitive cardiac troponinI: Unit: ng/L. Reference range: less than 14.
- hemoglobin: Unit: g/L. Reference range: 140 - 180 for men, 120 - 160 for women.
- Serum chloride: Unit: mmol/L. Reference range: 96 - 106.
- Prothrombin time: Unit: seconds. Reference range: 13.1 - 14.125.
- procalcitonin: Unit: ng/mL. Reference range: less than 0.05.
- eosinophils(%): Unit: %. Reference range: 1 - 6.
- Interleukin 2 receptor: Unit: pg/mL. Reference range: less than 625.
- Alkaline phosphatase: Unit: IU/L. Reference range: 44 - 147.
- albumin: Unit: g/dL. Reference range: 3.5 - 5.5.
- basophil(%): Unit: %. Reference range: 0.5 - 1.
- Interleukin 10: Unit: pg/mL. Reference range: less than 9.8.
- Total bilirubin: Unit: µmol/L. Reference range: 5.1 - 17.
- Platelet count: Unit: x 10^9/L. Reference range: 150 - 450.
- monocytes(%): Unit: %. Reference range: 2 - 10.
- antithrombin: Unit: %. Reference range: 80 - 120.
- Interleukin 8: Unit: pg/mL. Reference range: less than 62.
- indirect bilirubin: Unit: μmol/L. Reference range: 3.4 - 12.0.
- Red blood cell distribution width: Unit: %. Reference range: 11.5 - 14.5 for men, 
12.2 - 16.1 for women.
- neutrophils(%): Unit: %. Reference range: 45 - 70.
- total protein: Unit: g/L. Reference range: 60 - 83.
- Quantification of Treponema pallidum antibodies: Unit: /. Reference range: less than 
1.0.
- Prothrombin activity: Unit: %. Reference range: 70 - 130.
- HBsAg: Unit: IU/mL. Reference range: 0.0 - 0.01.
- mean corpuscular volume: Unit: fL. Reference range: 80 - 100.
- hematocrit: Unit: %. Reference range: 40 - 54 for men, 36 - 48 for women.
- White blood cell count: Unit: x 10^9/L. Reference range: 4.5 - 11.0.
- Tumor necrosis factorα: Unit: pg/mL. Reference range: less than 8.1.
- mean corpuscular hemoglobin concentration: Unit: g/L. Reference range: 320 - 360.
- fibrinogen: Unit: g/L. Reference range: 2 - 4.
- Interleukin 1β: Unit: pg/mL. Reference range: less than 6.5.
- Urea: Unit: mmol/L. Reference range: 1.8 - 7.1.
- lymphocyte count: Unit: x 10^9/L. Reference range: 1.0 - 4.8.
- PH value: Unit: /. Reference range: 7.35 - 7.45.
- Red blood cell count: Unit: x 10^12/L. Reference range: 4.5 - 5.5 for men, 4.0 - 5.0 
for women.
- Eosinophil count: Unit: x 10^9/L. Reference range: 0.02 - 0.5.
- Corrected calcium: Unit: mmol/L. Reference range: 2.12 - 2.57.
- Serum potassium: Unit: mmol/L. Reference range: 3.5 - 5.0.
- glucose: Unit: mmol/L. Reference range: 3.9 - 5.6.
- neutrophils count: Unit: x 10^9/L. Reference range: 2.0 - 8.0.
- Direct bilirubin: Unit: µmol/L. Reference range: 1.7 - 5.1.
- Mean platelet volume: Unit: fL. Reference range: 7.4 - 11.4.
- ferritin: Unit: ng/mL. Reference range: 24 - 336 for men, 11 - 307 for women.
- RBC distribution width SD: Unit: fL. Reference range: 40.0 - 55.0.
- Thrombin time: Unit: seconds. Reference range: 12 - 19.
- (%)lymphocyte: Unit: %. Reference range: 20 - 40.
- HCV antibody quantification: Unit: IU/mL. Reference range: 0.04 - 0.08.
- DD dimer: Unit: mg/L. Reference range: 0 - 0.5.
- Total cholesterol: Unit: mmol/L. Reference range: less than 5.17.
- aspartate aminotransferase: Unit: U/L. Reference range: 8 - 33.
- Uric acid: Unit: µmol/L. Reference range: 240 - 510 for men, 160 - 430 for women.
- HCO3: Unit: mmol/L. Reference range: 22 - 29.
- calcium: Unit: mmol/L. Reference range: 2.13 - 2.55.
- Aminoterminal brain natriuretic peptide precursor(NTproBNP): Unit: pg/mL. 
Reference range: 0 - 125.
- Lactate dehydrogenase: Unit: U/L. Reference range: 140 - 280.
- platelet large cell ratio: Unit: %. Reference range: 15 - 35.
- Interleukin 6: Unit: pg/mL. Reference range: 0 - 7.
- Fibrin degradation products: Unit: μg/mL. Reference range: 0 - 10.
- monocytes count: Unit: x 10^9/L. Reference range: 0.32 - 0.58.
- PLT distribution width: Unit: fL. Reference range: 9.2 - 16.7.
- globulin: Unit: g/L. Reference range: 23 - 35.
- γglutamyl transpeptidase: Unit: U/L. Reference range: 7 - 47 for men, 5 - 25 for women.
- International standard ratio: Unit: ratio. Reference range: 0.8 - 1.2.
- basophil count(#): Unit: x 10^9/L. Reference range: 0.01 - 0.02.
- mean corpuscular hemoglobin: Unit: pg. Reference range: 27 - 31.
- Activation of partial thromboplastin time: Unit: seconds. Reference range: 22 - 35.
- High sensitivity Creactive protein: Unit: mg/L. Reference range: 3 - 10.
- HIV antibody quantification: Unit: IU/mL. Reference range: 0.08 - 0.11.
- serum sodium: Unit: mmol/L. Reference range: 135 - 145.
- thrombocytocrit: Unit: %. Reference range: 0.22 - 0.24.
- ESR: Unit: mm/hr. Reference range: less than 15  for men, less than 20 for women.
- glutamicpyruvic transaminase: Unit: U/L. Reference range: 0 - 35.
- eGFR: Unit: mL/min/1.73m². Reference range: more than 90.
- creatinine: Unit: µmol/L. Reference range: 61.9 - 114.9 for men, 53 - 97.2 for women.

Here is an example of input information:
Example #1:
Input information of a patient:
The patient is a male, aged 52.0 years.
The patient had 5 visits that occurred at 2020-02-09, 2020-02-10, 2020-02-13, 
2020-02-14, 2020-02-17.
Details of the features for each visit are as follows:
- Hypersensitive cardiac troponinI: "1.9, 1.9, 1.9, 1.9, 1.9"
- hemoglobin: "139.0, 139.0, 142.0, 142.0, 142.0"
- Serum chloride: "103.7, 103.7, 104.2, 104.2, 104.2"
......

RESPONSE:
0.25

Example #2:
Input information of a patient:
The patient is a female, aged 71.0 years.
The patient had 5 visits that occurred at 2020-02-01, 2020-02-02, 2020-02-09, 
2020-02-10, 2020-02-11.
Details of the features for each visit are as follows:
- Hypersensitive cardiac troponinI: "5691.05, 11970.22, 9029.88, 6371.5, 3638.55"
- hemoglobin: "105.68, 132.84, 54.19, 136.33, 123.69"
- Serum chloride: "89.18, 101.54, 90.35, 103.99, 102.06"
......

RESPONSE:
0.85

Input information of a patient:
The patient is a male, aged 73.0 years.
The patient had 7 visits that occurred at 2020-01-31, 2020-02-04, 2020-02-06, 
2020-02-10, 2020-02-15, 2020-02-16, 2020-02-17.
Details of the features for each visit are as follows:
- Hypersensitive cardiac troponinI: "19.9, 19.9, 19.9, 19.9, 19.9, 19.9, 19.9"
- hemoglobin: "136.0, 136.0, 140.0, 130.0, 129.0, 131.0, 131.0"
- Serum chloride: "103.1, 103.1, 101.4, 98.5, 98.1, 100.0, 100.0"
- Prothrombin time: "13.9, 13.9, 13.9, 14.1, 14.1, 12.4, 12.4"
- procalcitonin: "0.09, 0.09, 0.09, 0.09, 0.09, 0.09, 0.09"
- eosinophils(%): "0.6, 0.6, 0.3, 0.2, 1.1, 1.7, 1.7"
- Interleukin 2 receptor: "722.0, 722.0, 722.0, 722.0, 722.0, 722.0, 722.0"
- Alkaline phosphatase: "46.0, 46.0, 54.0, 57.0, 61.0, 71.0, 71.0"
- albumin: "33.3, 33.3, 33.2, 32.4, 35.9, 37.6, 37.6"
- basophil(%): "0.3, 0.3, 0.1, 0.1, 0.3, 0.2, 0.2"
- Interleukin 10: "9.9, 9.9, 9.9, 9.9, 9.9, 9.9, 9.9"
- Total bilirubin: "8.3, 8.3, 7.4, 16.6, 9.6, 6.3, 6.3"
- Platelet count: "105.0, 105.0, 214.0, 168.0, 143.0, 141.0, 141.0"
- monocytes(%): "10.7, 10.7, 7.2, 4.9, 9.0, 7.9, 7.9"
- antithrombin: "84.5, 84.5, 84.5, 84.5, 84.5, 84.5, 84.5"
- Interleukin 8: "17.6, 17.6, 17.6, 17.6, 17.6, 17.6, 17.6"
- indirect bilirubin: "4.3, 4.3, 4.5, 11.1, 6.0, 3.7, 3.7"
- Red blood cell distribution width : "11.9, 11.9, 11.6, 11.9, 11.9, 11.9, 11.9"
- neutrophils(%): "65.8, 65.8, 66.5, 84.3, 60.9, 64.3, 64.3"
- total protein: "69.3, 69.3, 67.9, 62.2, 67.2, 67.7, 67.7"
- Quantification of Treponema pallidum antibodies: "0.05, 0.05, 0.05, 0.05, 0.05, 0.05, 
0.05"
- Prothrombin activity: "91.0, 91.0, 91.0, 89.0, 89.0, 115.0, 115.0"
- HBsAg: "0.03, 0.03, 0.03, 0.03, 0.03, 0.03, 0.03"
- mean corpuscular volume: "91.8, 91.8, 91.1, 92.7, 93.2, 93.8, 93.8"
- hematocrit: "39.2, 39.2, 39.7, 38.0, 36.9, 38.0, 38.0"
- White blood cell count: "3.5700000000000003, 3.5700000000000003, 6.9, 12.58, 9.05, 
9.67, 9.67"
- Tumor necrosis factorα: "8.8, 8.8, 8.8, 8.8, 8.8, 8.8, 8.8"
- mean corpuscular hemoglobin concentration: "347.0, 347.0, 353.0, 342.0, 350.0, 345.0, 
345.0"
- fibrinogen: "4.41, 4.41, 4.41, 3.28, 3.28, 3.16, 3.16"
- Interleukin 1β: "6.9, 6.9, 6.9, 6.9, 6.9, 6.9, 6.9"
- Urea: "8.5, 8.5, 5.0, 7.6, 6.9, 6.5, 6.5"
- lymphocyte count: "0.8, 0.8, 1.79, 1.32, 2.6, 2.5, 2.5"
- PH value: "6.7075, 6.7075, 6.7075, 6.7075, 6.7075, 6.7075, 6.7075"
- Red blood cell count: "2.9349999999999996, 2.9349999999999996, 4.36, 4.1, 3.96, 4.05, 
4.05"
- Eosinophil count: "0.02, 0.02, 0.02, 0.02, 0.1, 0.16, 0.16"
- Corrected calcium: "2.29, 2.29, 2.53, 2.33, 2.47, 2.44, 2.44"
- Serum potassium: "4.33, 4.33, 4.73, 4.21, 4.61, 5.15, 5.15"
- glucose: "7.35, 7.35, 5.92, 17.18, 6.44, 6.75, 6.75"
- neutrophils count: "2.33, 2.33, 4.58, 10.61, 5.51, 6.23, 6.23"
- Direct bilirubin: "4.0, 4.0, 2.9, 5.5, 3.6, 2.6, 2.6"
- Mean platelet volume: "11.9, 11.9, 10.9, 10.5, 11.5, 11.3, 11.3"
- ferritin: "675.6, 675.6, 675.6, 675.6, 675.6, 634.9, 634.9"
- RBC distribution width SD: "40.8, 40.8, 39.0, 40.5, 40.7, 41.5, 41.5"
- Thrombin time: "16.9, 16.9, 16.9, 19.2, 19.2, 16.3, 16.3"
- (%)lymphocyte: "22.6, 22.6, 25.9, 10.5, 28.7, 25.9, 25.9"
- HCV antibody quantification: "0.06, 0.06, 0.06, 0.06, 0.06, 0.06, 0.06"
- D-D dimer: "2.2, 2.2, 2.2, 0.66, 0.66, 0.92, 0.92"
- Total cholesterol: "3.9, 3.9, 3.81, 3.65, 4.62, 4.84, 4.84"
- aspartate aminotransferase: "33.0, 33.0, 35.0, 16.0, 21.0, 23.0, 23.0"
- Uric acid: "418.0, 418.0, 281.0, 379.0, 388.0, 376.0, 376.0"
- HCO3-: "21.2, 21.2, 26.7, 25.6, 31.0, 28.0, 28.0"
- calcium: "2.02, 2.02, 2.25, 2.04, 2.25, 2.25, 2.25"
- Amino-terminal brain natriuretic peptide precursor(NT-proBNP): "60.0, 60.0, 60.0, 
60.0, 60.0, 60.0, 60.0"
- Lactate dehydrogenase: "306.0, 306.0, 250.0, 200.0, 198.0, 206.0, 206.0"
- platelet large cell ratio : "39.9, 39.9, 32.1, 29.3, 37.2, 36.9, 36.9"
- Interleukin 6: "26.064999999999998, 26.064999999999998, 26.064999999999998, 
26.064999999999998, 26.064999999999998, 26.064999999999998, 26.064999999999998"
- Fibrin degradation products: "17.65, 17.65, 17.65, 17.65, 17.65, 17.65, 17.65"
- monocytes count: "0.38, 0.38, 0.5, 0.62, 0.81, 0.76, 0.76"
- PLT distribution width: "16.3, 16.3, 12.6, 11.9, 14.9, 14.3, 14.3"
- globulin: "36.0, 36.0, 34.7, 29.8, 31.3, 30.1, 30.1"
- γ-glutamyl transpeptidase: "24.0, 24.0, 31.0, 27.0, 42.0, 41.0, 41.0"
- International standard ratio: "1.06, 1.06, 1.06, 1.08, 1.08, 0.92, 0.92"
- basophil count(#): "0.01, 0.01, 0.01, 0.01, 0.03, 0.02, 0.02"
- mean corpuscular hemoglobin : "31.9, 31.9, 32.1, 31.7, 32.6, 32.3, 32.3"
- Activation of partial thromboplastin time: "39.0, 39.0, 39.0, 37.9, 37.9, 38.9, 38.9"
- Hypersensitive c-reactive protein: "43.1, 43.1, 3.6, 3.6, 2.6, 2.6, 2.6"
- HIV antibody quantification: "0.09, 0.09, 0.09, 0.09, 0.09, 0.09, 0.09"
- serum sodium: "137.7, 137.7, 142.9, 139.4, 140.0, 142.7, 142.7"
- thrombocytocrit: "0.12, 0.12, 0.23, 0.18, 0.16, 0.16, 0.16"
- ESR: "41.0, 41.0, 41.0, 41.0, 41.0, 41.0, 41.0"
- glutamic-pyruvic transaminase: "16.0, 16.0, 42.0, 29.0, 29.0, 30.0, 30.0"
- eGFR: "46.6, 46.6, 72.7, 64.8, 74.7, 74.7, 74.7"
- creatinine: "130.0, 130.0, 90.0, 99.0, 88.0, 88.0, 88.0"

Please respond with only a floating-point number between 0 and 1, where a higher number 
suggests a greater likelihood of death. Do not include any additional explanation.
RESPONSE:
\end{Verbatim}
% \end{BVerbatim}
% \end{verbatim}
\end{tcolorbox}
    \item \textbf{MIMIC-IV Dataset Mortality Prediction Task:} 
% \begin{lstlisting}
% \end{lstlisting}

\begin{tcolorbox}[title=Complete Prompt Template for Optimal Settings in the TJH Dataset,
    colback=white,
    colframe=black,
    colbacktitle=white,
    coltitle=black,
    fonttitle=\bfseries,
    breakable,
    enhanced]
% \begin{verbatim}
% \begin{BVerbatim}[fontsize=\small]
\begin{Verbatim}[fontsize=\small]
You are an experienced doctor in Intensive Care Unit (ICU) treatment.

I will provide you with medical information from multiple Intensive Care Unit (ICU) 
visits of a patient, each characterized by a fixed number of features.

Present multiple visit data of a patient in one batch. Represent each feature within 
this data as a string of values, separated by commas.

Your task is to assess the provided medical data and analyze the health records from 
ICU visits to determine the likelihood of the patient not surviving their hospital 
stay. 
Please respond with only a floating-point number between 0 and 1, where a higher 
number suggests a greater likelihood of death.

In situations where the data does not allow for a reasonable conclusion, respond with 
the phrase `I do not know` without any additional explanation.

- Capillary refill rate: Unit: /. Reference range: /.
- Glascow coma scale eye opening: Unit: /. Reference range: /.
- Glascow coma scale motor response: Unit: /. Reference range: /.
- Glascow coma scale total: Unit: /. Reference range: /.
- Glascow coma scale verbal response: Unit: /. Reference range: /.
- Diastolic blood pressure: Unit: mmHg. Reference range: less than 80.
- Fraction inspired oxygen: Unit: /. Reference range: more than 0.21.
- Glucose: Unit: mg/dL. Reference range: 70 - 100.
- Heart Rate: Unit: bpm. Reference range: 60 - 100.
- Height: Unit: cm. Reference range: /.
- Mean blood pressure: Unit: mmHg. Reference range: less than 100.
- Oxygen saturation: Unit: %. Reference range: 95 - 100.
- Respiratory rate: Unit: breaths per minute. Reference range: 15 - 18.
- Systolic blood pressure: Unit: mmHg. Reference range: less than 120.
- Temperature: Unit: degrees Celsius. Reference range: 36.1 - 37.2.
- Weight: Unit: kg. Reference range: /.
- pH: Unit: /. Reference range: 7.35 - 7.45.

Here is an example of input information:
Example #1:
Input information of a patient:
The patient is a female, aged 52 years.
The patient had 4 visits that occurred at 0, 1, 2, 3.
Details of the features for each visit are as follows:
- Capillary refill rate: "unknown, unknown, unknown, unknown"
- Glascow coma scale eye opening: "Spontaneously, Spontaneously, Spontaneously, 
Spontaneously"
- Glascow coma scale motor response: "Obeys Commands, Obeys Commands, Obeys Commands, 
Obeys Commands"
...

RESPONSE:
0.3

Input information of a patient:
The patient is a male, aged 50.0 years.
The patient had 4 visits that occurred at 0, 1, 2, 3.
Details of the features for each visit are as follows:
- Capillary refill rate: "unknown, unknown, unknown, unknown"
- Glascow coma scale eye opening: "None, None, None, None"
- Glascow coma scale motor response: "Flex-withdraws, Flex-withdraws, unknown, 
Localizes Pain"
- Glascow coma scale total: "unknown, unknown, unknown, unknown"
- Glascow coma scale verbal response: "No Response-ETT, No Response-ETT, 
No Response-ETT, No Response-ETT"
- Diastolic blood pressure: "79.41666666666667, 77.83333333333333, 85.83333333333333, 
83.25"
- Fraction inspired oxygen: "0.5, 0.5, 0.5, 0.5"
- Glucose: "172.0, 150.0, 128.0, 147.0"
- Heart Rate: "85.41666666666667, 84.91666666666667, 87.33333333333333, 
88.41666666666667"
- Height: "173.0, 173.0, 173.0, 173.0"
- Mean blood pressure: "96.41666666666667, 97.58333333333333, 109.91666666666667, 
108.41666666666667"
- Oxygen saturation: "99.5, 100.0, 100.0, 100.0"
- Respiratory rate: "22.083333333333332, 21.583333333333332, 22.0, 22.25"
- Systolic blood pressure: "136.5, 135.41666666666666, 152.16666666666666, 
153.16666666666666"
- Temperature: "37.31944444444444, 37.074074074074076, 37.36296296296297, 
37.76666666666667"
- Weight: "69.4, 69.4, 68.33174919999999, 68.2"
- pH: "7.484999999999999, 7.484999999999999, 7.45, 7.43"

Please respond with only a floating-point number between 0 and 1, where a higher number 
suggests a greater likelihood of death. Do not include any additional explanation.
RESPONSE:

\end{Verbatim}
% \end{BVerbatim}
% \end{verbatim}
\end{tcolorbox}
\end{itemize}

\end{document}